\documentclass[10pt,twocolumn,letterpaper]{article}

\usepackage[pagenumbers]{wacv} 

\definecolor{wacvblue}{rgb}{0.21,0.49,0.74}
\usepackage[pagebackref,breaklinks,colorlinks,allcolors=wacvblue]{hyperref}

\usepackage{multirow}
\usepackage[acronym]{glossaries}
\usepackage{dsfont}
\usepackage{tabularx}
\usepackage{booktabs}
\usepackage{array}
\newacronym{cv}{CV}{Computer Vision}
\newacronym{dlt}{DLT}{Direct Linear Transformation}
\newacronym{uq}{UQ}{Uncertainty Quantification}
\newacronym{he}{HE}{homography estimation}
\newacronym{rmse}{RMSE}{Root Mean Squared Error}
\newacronym{sift}{SIFT}{Scale Invariant Feature Transform}
\newacronym{bi}{BI}{Bayesian Inference}
\newacronym{ekf}{EKF}{Extended Kalman Filter}
\newacronym{imm}{IMM}{Interacting Multiple Model}
\newacronym{cc}{CC}{Camera Calibration}
\newacronym{mcmc}{MCMC}{Markov Chain Monte Carlo}
\newacronym{poc}{PoC}{Proof of Concept}

\usepackage{xr}
\title{Closed-form Bayesian homography estimation from noisy point correspondences}

\author{Hanne Beuter\\
Technische Hochschule Augsburg\\
TTZ Landsberg am Lech\\
Augsburg, Germany \\
{\tt\small hanne.beuter@tha.de}
\and
Sebastian Dorn\\
Technische Hochschule Augsburg\\
TTZ Landsberg am Lech\\
Augsburg, Germany \\
{\tt\small sebastian.dorn@tha.de}
}

\begin{document}
\maketitle
\begin{abstract}
While homographies are fundamental to many computer vision tasks, the majority of conventional estimation techniques provide only point estimates without directly quantifying uncertainty introduced by noisy observations. Uncertainty, though, propagates to subsequent processing steps such as camera calibration and 3D reconstruction and is particularly relevant in safety-critical and socially relevant fields including medical imaging, autonomous driving, and defense.
We present a fast Bayesian formulation for homography estimation from point correspondences that explicitly incorporates measurement uncertainty and prior knowledge while providing a posterior distribution over the homography parameters. A closed-form solution of the posterior mean of the homography is derived in homogeneous coordinates and supplemented by an iterative Bayesian approach to handle non-linearities. Synthetic experiments demonstrate the applicability to projective transformations and show improved estimation accuracy over DLT under varying noise conditions. Image stitching experiments further demonstrate applicability to real image correspondences while additionally providing uncertainty information.
\end{abstract}    
\section{Introduction}
\label{sec:intro}
A homography is a projective transformation that maps the corresponding points between two images of the same planar scene taken from different viewpoints \cite{szeliskiComputerVisionAlgorithms2022, hartleyMultipleViewGeometry2004} and is a central concept in \gls*{cv}. Many fundamental methods, including image registration and stitching \cite{szeliskiImageAlignmentStitching2007}, Bird's Eye View transformations \cite{bertozziStereoInversePerspective1998}, \gls*{cc} \cite{zhangFlexibleNewTechnique2000}, and visual SLAM \cite{murorbslam2015}, rely on it. 
Medical imaging \cite{atanasijevicOpenSourceApplicationMri2022, mundadaAutomatedXRayImage2025}, UAV geolocation in defense \cite{baykalImageMatchingUAV2025}, and autonomous driving \cite{bertozziStereoInversePerspective1998,yooroaduserposition2026} are examples of safety-critical application fields utilizing homographies where errors in perception can directly affect system decisions or outputs and consequently pose a risk to human safety \cite{yooroaduserposition2026,songNOVELITERATIVEMATCHING2020}. This makes the uncertainty of an estimated homography matrix important information for subsequent processing in homography-based \gls*{cv} systems.
Traditional \gls*{he} methods such as \gls*{dlt} \cite{hartleyMultipleViewGeometry2004} provide deterministic point estimates rather than directly inferring parameter uncertainty or incorporating prior parameter knowledge. In contrast, \gls*{bi} provides both by posterior inference \cite{gelmanBayesianDataAnalysis2014}. However, recent implementations require computationally expensive sampling \cite{sundareswara_bayesian_2005, gutierrez-moizant_novel_2023} or information about camera motion over time \cite{bernal_bayesian_2023, claasenVideobasedSequentialBayesian2024}.
We derive a closed-form Bayesian inference method including \gls*{uq} for feature-based \gls*{he} on static image correspondences. Our contributions are as follows:

\begin{itemize}
    \item Theoretical derivation of the inference method to determine the posterior mean of a homography from point correspondences.
    \item \gls*{uq} of the homography matrix parameters via posterior distributions.
    \item \gls*{poc} on synthetic and image stitching data with application specific noise modeling. 
\end{itemize}

\section{Related work}
\label{sec:relatedwork}
In order to determine the homography transformations, feature detectors usually first identify keypoints and assign them pairwise, before the homographic parameters are estimated by standard methods of linear algebra, e.g. by the \gls*{dlt} \cite{hartleyMultipleViewGeometry2004}. The robustness of the solution is usually ensured by removing outlier keypoints with additional algorithms like RANSAC \cite{fischlerRandomSampleConsensus1981}, resulting in precise estimates for accurately localized point correspondences. Estimation accuracy, however, decreases as localization noise increases \cite{zhaoAccurateRobustFeaturebased2016a}.
Although often precise enough, the homography matrix remains a point estimate without directly providing information on its uncertainty \cite{raguramExploitingUncertaintyRandom2009}. While likelihood-based error propagation is used \cite{hartleyMultipleViewGeometry2004}, standard \gls*{dlt}-based estimation does not allow for the inclusion of prior knowledge potentially available from previous experiments or expert judgments. 
\gls*{bi} allows treating unknown parameters as random variables and inferring posterior distributions while explicitly modeling uncertainty \cite{gelmanBayesianDataAnalysis2014}. Bayesian approaches for \gls*{he} and \gls*{cc} share the central motivation of mitigating the consequences of uncertain parameter estimates for subsequent processing \cite{bernal_bayesian_2023, claasenVideobasedSequentialBayesian2024, gutierrez-moizant_novel_2023, sundareswara_bayesian_2005}.
Previous works \cite{bernal_bayesian_2023, claasenVideobasedSequentialBayesian2024} formulate sequential Bayesian filtering frameworks to exploit information from preceding video frames for \gls*{he}. In \cite{bernal_bayesian_2023}, visual feature correspondences, rate-gyro measurements and a camera motion model are combined within a Bayesian state estimator, providing both a homography estimate and an associated state covariance. The approach in \cite{claasenVideobasedSequentialBayesian2024} instead derives information about camera movement entirely from image observations. A two-stage Kalman filter first tracks keypoint positions and their uncertainties and subsequently estimates the homography and its uncertainty while incorporating information from preceding video frames.
Both approaches improve robustness when visual information is limited, but rely on sequential observations and temporal information.

Bayesian estimation without a temporal component has also been investigated for camera parameter estimation \cite{sundareswara_bayesian_2005, gutierrez-moizant_novel_2023}. The method in \cite{sundareswara_bayesian_2005} formulates a posterior over extrinsic camera parameters based on noisy point correspondences, allowing the resulting parameter uncertainty to be propagated into subsequent 3D reconstruction. Another method \cite{gutierrez-moizant_novel_2023} considers uncertainty characterization for Zhang's camera calibration procedure \cite{zhangFlexibleNewTechnique2000} and infers intrinsic and lens-distortion parameters. Although homographies are used for the estimation of initial extrinsic parameters within the calibration procedure, they are not themselves inferred probabilistically. A common computational aspect of these non-temporal approaches is that the resulting posterior distributions are not evaluated in closed form and computationally expensive \gls*{mcmc} sampling is used \cite{sundareswara_bayesian_2005, gutierrez-moizant_novel_2023}. 

Taken together, the related methods demonstrate the effectiveness of Bayesian methods for uncertainty-aware geometric vision problems, ranging from \gls*{he} to \gls*{cc}. However, the approaches are either limited by a necessity for sampling techniques, leading to approximations and high run-time, making the methods not real-time applicable, or their dependence on video sequences for temporal information. In contrast to these approaches, our work introduces an exact, closed-form solution to infer homographic projections. It can cope with noisy observations (e.g. from feature detections), few data points and is computationally cheap. Additionally, it allows for incorporation of prior knowledge.

\section{Methodology}
\label{sec:methodology}
We aim to introduce Bayesian inference theory to estimate homographic transformations especially for 2D image points in 3D homogeneous coordinates. Since all findings below hold without any restrictions for arbitrarily high dimensions, we carry out the derivations for the general case of points in $k-$dim. homogeneous coordinates. 

\subsection{Observation model}
\label{subsec:observationmodel}
We observe a given image $\mathcal{S}$ of a specific scenery in $(k-1)$ dimensions. E.g. for $k=3$ we observe a 2D image as a projection of a 3D scenery. A point within $\mathcal{S}$ in homogeneous coordinates is given by $s_i \in \mathbb{P}^{k-1} \subset \mathbb{R}^k$.
The specific points, following named keypoints, might have been detected automatically by a computer vision algorithm. Subsequently, we observe a second image $\mathcal{D}$ including parts of the scenery observed already in $\mathcal{S}$ from a different perspective. The relation between the image points is modeled as follows: Out of all points in $\mathcal{S}$, a specific subset of $\mathrm{n}$ points $s_i, ~i=1,..., \mathrm{n}$, is respectively transformed by a homographic operator $\mathbf{R}\in\mathbb{R}^{k\times k}$ to new data points $d_i$ being part of the second image $\mathcal{D}$. The data generation is in general affected by some noise $n_i$, that is in general different for each point $d_i$. This might be caused in practice by, e.g., non-optimal keypoints detection/location or deviations from homographic assumptions (planar scenes). Subsequently the noise is considered to be additive and Gaussian\footnote{The derivation can be repeated with any other noise model.},  ${n}_i \sim \mathcal{G}(0, \mathbf{N}_i)$ with
\begin{equation}
    \mathcal{G}(n_i, \mathbf{N}_i)=|2\pi \mathbf{N}_i|^{-\frac{1}{2}}\exp{\left(-\frac{1}{2}n_i^t \mathbf{N}_i^{-1} n_i\right)},
\label{eq:gaussnoise}
\end{equation}
where $\mathbf{N}_i$ denotes the noise covariance matrix, $|\cdot|$ the determinant, and $(\cdot)^t$ the transposition operation. The observation equation is then given by
\begin{equation}
    d_i = \mathbf{R} {s}_i + {n}_i.
\end{equation}
In vector notation the equation reads $d = \hat{\mathbf{R}} {s} + {n}$,

\begin{equation}
\hat{\mathbf{R}} =
\begin{bmatrix}
    \textbf{R} & 0  & \cdots & 0 \\
    0 & \textbf{R}  & \cdots & 0 \\
    \vdots & \vdots & \ddots & \vdots \\
    0 & 0 & \cdots & \textbf{R}
\end{bmatrix}
, ~\hat{\mathbf{N}} =
\begin{bmatrix}
    \textbf{N}_1 & 0  & \cdots & 0 \\
    0 & \textbf{N}_2  & \cdots & 0 \\
    \vdots & \vdots & \ddots & \vdots \\
    0 & 0 & \cdots & \textbf{N}_\mathrm{n}
\end{bmatrix}~~,
\end{equation}
with $s,~n,~d\in\mathbb{R}^{k\mathrm{n}}$ and $\hat{\mathbf{R}}, \hat{\mathbf{N}}\in\mathbb{R}^{k\mathrm{n}\times k\mathrm{n}}$. Note that also the original points $s_i$ might suffer from noise. Due to the linear structure of the observation equation, this can be absorbed in $n_i$ as well. 

The subsequent general-purpose derivation is done using a generic Gaussian noise model without further restrictions to $\mathbf{N}_i$. Specific noise models that are relevant for \gls{cv} are discussed in \cref{sec:noisemodel}. 

\subsection{Prior knowledge about the homographic transformation}
The goal, as described in \cref{sec:intro}, is to statistically infer the homographic transformation $\mathbf{R}$, given ${d}$ and ${s}$ (observed images). This means that we treat $\mathbf{R}$ as a statistical quantity with statistical properties like mean and covariance. In order to infer $\mathbf{R}$, we calculate the posterior probability distribution
\begin{align}
    \mathcal{P}(\mathbf{R} | d, {s}) =& ~\frac{\mathcal{P}({d},s | \mathbf{R}) \mathcal{P}(\mathbf{R})}{\mathcal{P}({d}, \mathbf{s})} = \frac{\mathcal{P}(d|\mathbf{R},s)\mathcal{P}(\mathbf{R})}{\mathcal{P}(d|s)}\\
    \propto & ~\mathcal{P}(d|\mathbf{R},s)\mathcal{P}(\mathbf{R}). 
    \label{eq:bayes}
\end{align}
The prior term $\mathcal{P}(\mathbf{R})$ in equation (\ref{eq:bayes}) enables us to include prior knowledge about the homographic transformation itself. If, for instance, the type of transformation is already known (euclidean, affine, \ldots) and/or the expected range of the parameters, describing the transformation, can be estimated, this information can be included in $\mathcal{P}(\mathbf{R})$. 

To model the prior, we choose within this work a Gaussian probability distribution. Reasons for this choice rely theoretically on maximum entropy arguments for potentially known first and second moment of the distribution as well as practically on the fact that a Gaussian is a conjugate prior to the likelihood of equation (\ref{eq:bayes}), i.e. the posterior is a Gaussian as well. Since $\mathbf{R}$ is a matrix, a matrix normal distribution is required,
\begin{align}
    &\mathcal{P}(\mathbf{R})= \mathcal{G}(\mathbf{R}, (\mathbf{U},\mathbf{V}))\\
    =&(2\pi)^{-\frac{k^2}{2}}|\mathbf{V}|^{-\frac{k}{2}} |\mathbf{U}|^{-\frac{k}{2}}\mathrm{e}^{\left(-\frac{1}{2}\mathrm{tr}\left[\mathbf{V}^{-1}\mathbf{R}^t \mathbf{U}^{-1}\mathbf{R} \right]\right)} \label{eq:prior}\\
    =& (2\pi)^{-\frac{k^2}{2}} |\mathbf{V}\otimes\mathbf{U}|^{-\frac{1}{2}} \mathrm{e}^{\left(-\frac{1}{2}\mathrm{vec}(\mathbf{R})^t (\mathbf{V}\otimes \mathbf{U})^{-1}\mathrm{vec}(\mathbf{R}) \right)}\\
    =&\mathcal{G}(\mathrm{vec}(\mathbf{R}),\mathbf{V}\otimes\mathbf{U})~~, \label{eq:prior_vec}
\end{align}
with covariance matrices $\mathbf{U}, \mathbf{V}\in\mathbb{R}^{k\times k}$ to model the correlation structure among the rows and columns, respectively. $\mathrm{vec(\cdot)}$ denotes the vectorization of a matrix and $\otimes$ the Kronecker product.
The above formulation assumes a zero-mean prior, but prior knowledge of an expected homography $\mathbf{R}_0$ can be incorporated as
$\mathcal{P}(\mathbf{R}) = \mathcal{G}(\mathrm{vec}(\mathbf{R})-\mathrm{vec}(\mathbf{R}_0), \mathbf{V} \otimes \mathbf{U})$.

In fact, a homographic mapping, as described in homogeneous coordinates, only has $8$ degrees of freedom. In practice, however, it is common to infer all components of $\mathbf{R}$ followed by a normalization, e.g. by setting $\mathbf{R}[9]=1$, or $||\mathrm{vec}(\mathbf{R})||_2^2=1$, etc.

\subsection{Bayesian inference of the homographic transformation}
\label{subsec:bayesianofthehomographic}
To perform the Bayesian inference step we need the likelihood,
\begin{align}
    &\mathcal{P}(d|\mathbf{R},s) = \int dn \mathcal{P}(d|\mathbf{R}s, n)\mathcal{P}(n|\mathbf{R},s)\\
    &= \int \delta(d-\hat{\mathbf{R}}s-n)\mathcal{G}(n,N)=\mathcal{G}(d-\hat{\mathbf{R}}s,N). \\
    &\equiv \exp\left({-\mathcal{H}(d|\mathbf{R},s)}\right)
    \label{eq:likelihood}
\end{align} 
where the information $\mathcal{H}$ was introduced, 
\begin{align}
    \mathcal{H}(\mathbf{R} | d, {s})=&~-\ln \mathcal{P}(\mathbf{R} | d, {s})\\
    =&~-\ln\mathcal{P}(d|\mathbf{R},s)-\ln\mathcal{P}(\mathbf{R})+\mathcal{H}_0.
\label{eq:surprise0}
\end{align}
$\mathcal{H}_0$ (and $\mathcal{H}'_0$ below) summarize all terms constant in $\mathbf{R}$. Inserting \cref{eq:prior,eq:likelihood} into \cref{eq:surprise0} yields the posterior information
\begin{align}
    \mathcal{H}(\mathbf{R} | d, {s}&)=\frac{1}{2}s^t\hat{\mathbf{R}}^t\hat{\mathbf{N}}^{-1}\hat{\mathbf{R}}s-d^t\hat{\mathbf{N}}^{-1}\hat{\mathbf{R}}s\\ 
    &~~~~~+\frac{1}{2}\mathrm{tr}\left[\mathbf{V}^{-1}\mathbf{R}^t \mathbf{U}^{-1}\mathbf{R} \right] + \mathcal{H}'_0\\
    &=\frac{1}{2}\sum_i\left( s_i^t\mathbf{R}^t\mathbf{N}_i^{-1}\mathbf{R}s_i -2d_i^t \mathbf{N}_i^{-1}\mathbf{R}s_i\right) \\
    &~~~~+\frac{1}{2}\mathrm{tr}\left[\mathbf{V}^{-1}\mathbf{R}^t \mathbf{U}^{-1}\mathbf{R} \right] + \mathcal{H}'_0\\
    &\equiv\frac{1}{2}\sum_i\left[(\mathcal{H}_1)_i - (\mathcal{H}_2)_i\right] +\mathcal{H}_\mathrm{prior} + \mathcal{H}'_0.
\label{eq:surprise}
\end{align}
Given the posterior information we can calculate the \textit{maximum a posteriori} estimate $\mathbf{R}_\mathrm{MAP}$ by determining the global minima of $\mathcal{H}(\mathbf{R} | d, {s})$. Since the posterior information contains terms up to quadratic order in $\mathbf{R}$ the MAP solution is simultaneously the posterior mean solution $\mathbf{R}_\mathrm{m}$ 
\begin{equation}
    \mathbf{R}_{\mathbf{m},kl} = \int \left(\prod\nolimits_{k,l} d\mathbf{R}_{kl}\right)
\mathbf{R}_{kl}~ \mathcal{P}(\mathbf{R}|d,s).
\end{equation}
Thus, the following matrix derivatives are needed
\begin{equation}
    \frac{\partial(\mathcal{H}_1)_i}{\partial \mathbf{R}}=2\mathbf{N}_i^{-1}\mathbf{R}(s_is_i^t) ~~,
\end{equation}
\begin{equation}
    \frac{\partial(\mathcal{H}_2)_i}{\partial \mathbf{R}}=2\mathbf{N}_i^{-1} (d_i s_i^t)  ~~,\frac{\partial\mathcal{H}_\mathrm{prior}}{\partial \mathbf{R}}=\mathbf{U^{-1}RV^{-1}}  ~~.
\end{equation}
Taking the derivate of (\cref{eq:surprise}) with regard to $\mathbf{R}$, and equating to zero then leads to 
\begin{align}
    \sum_i\left[ \mathbf{N}_i^{-1}\mathbf{R}_\mathrm{m}(s_is_i^t)-\mathbf{N}_i^{-1} (d_i s_i^t)\right]+\mathbf{U}^{-1}\mathbf{R}_\mathrm{m} \mathbf{V^{-1}}=0.  
    \label{eq:map-full0}
\end{align}

Vectorization of \cref{eq:map-full0} and solving for $\mathbf{R_\mathrm{m}}$ yields
\begin{equation}
\begin{aligned}
    &\text{vec}(\mathbf{R_\mathrm{m}})=\\
    &\left[\sum_i \left(s_i s_i^t \right)\otimes \mathbf{N}_i^{-1} +(\mathbf{V}^t \otimes \mathbf{U})^{-1} \right]^{-1}\sum_i s_i \otimes \mathbf{N}_i^{-1}d_i ~, \label{eq:wienerf}
\end{aligned}
\end{equation}
and represents the closed-form Wiener filter \cite{wiener1949extrapolation} solution.

\subsection{Simplification of the observation model}
\label{subsec:simplificationobservation}
 \cref{eq:map-full0,eq:wienerf} provides a way to compute the posterior mean $\mathbf{R}_\mathrm{m}$ in general for arbitrary covariance matrices $\mathbf{U, V, N}_i$. In practice, it is often more convenient and even more realistic to apply the following simplifications: 
\begin{itemize}
    \item Noise: \textbf{a)} Instead of using individual Noise covariances $\mathbf{N}_i$ for each data point, one might assume that all points obey one and the same statistics, i.e. $\mathbf{N_i}=\mathbf{N} ~\forall i$. Often it is sufficient to use white noise statistics, $\mathbf{N}=\sigma_n^2 \mathds{1}^{\mathrm{k}\times\mathrm{k}}\equiv \mathrm{\mathbf{diag}}(\sigma_n^2)$.\\
    \textbf{b)} If it is required to have individual noise covariances per data point, e.g. due to local image properties affecting the keypoints detection algorithm, one could still use $\mathbf{N}_i= \mathrm{\mathbf{diag}}(\sigma_{n,i}^2)$.
    \item Prior: General covariance matrices $\mathbf{U, V}$ allow for arbitrary correlations among rows and columns. Often, however, it is sufficient to assume an isotropic scenario: $\mathbf{U}=\sigma^2_u \mathds{1}^{\mathrm{k}\times\mathrm{k}}\equiv \mathrm{\mathbf{diag}}(\sigma_u^2), \mathbf{V}=\sigma^2_v \mathds{1}^{\mathrm{k}\times\mathrm{k}}\equiv \mathrm{\mathbf{diag}}(\sigma_v^2) $. I.e. the entries of distribution samples are statistically independent and identical.   
\end{itemize}

Applying the prior and the noise simplifications \textbf{a} to Equation (\ref{eq:map-full0}) leads to 

\begin{align}
\mathbf{R}_\mathrm{m, a)}=&
    \left(\sum_i d_i s_i^t\right)\left(\mathbf{diag}\left(\frac{\sigma_n^{2}}{\sigma_v^{2}\sigma_u^{2}}\right) +\sum_i   
    s_is_i^t\right)^{-1},  
    \label{eq:map-full}
\end{align}
with $\mathbf{diag}(x)$ denoting a diagonal matrix with diagonal elements equal to the scalar $x$. Using instead noise simplification \textbf{b)} yields
\begin{align}
&\mathbf{R}_\mathrm{m,b)}=\left(\sum_i \mathbf{diag}\left(\sigma_{n,i}^{-2}\right) \left(d_i s_i^t\right)\right)\\
&\times\left(\mathbf{diag}\left(\sigma_v^{-2}\sigma_u^{-2}\right) +\sum_i \mathbf{diag}\left(\sigma_{n,i}^{-2}\right)  
    \left(s_is_i^t\right)\right)^{-1}.  
    \label{eq:map-full2}
\end{align}

Interpretation of equations (\ref{eq:map-full},~\ref{eq:map-full2}): Without any prior knowledge with regard to the transformation, i.e. $\sigma^{-2}_u=\sigma^{-2}_v =0$, the equations (\ref{eq:map-full},~\ref{eq:map-full2}) are consistent with findings in optimal linear filter theory for inference of scalar-valued quantities \cite{kailath1974}. \cref{eq:map-full2} reduces to
\begin{align}
    \mathbf{R}_\mathrm{m,b)}=&~ \left(\sum_i \mathbf{diag}\left(\sigma_{n,i}^{-2}\right) \left(d_i s_i^t\right)\right) \\
    &\times\left(\sum_i \mathbf{diag}\left(\sigma_{n,i}^{-2}\right)  
    \left(s_is_i^t\right)\right)^{-1}
    \label{eq:rmb_noprior}
\end{align}
that is the ratio between a noise-weighted sum over $d_is_i^t$ and $s_is_i^t$, respectively. Data points with huge noise, $\sigma^2_{n,i}~\approx \infty$ will not contribute to the sum. For $\sigma^2_{n,i}=\sigma^2_n ~\forall i$ the result becomes independent of noise
\begin{equation}
\mathbf{R}_\mathrm{m,b)}= \left(\sum_i \left(d_i s_i^t\right)\right)\left(\sum_i   
    \left(s_is_i^t\right)\right)^{-1}.    
    \label{eq:simple}
\end{equation}

\subsection{Uncertainty quantification}
\label{subsec:uq}
There are several fields of application where reliable results of computer vision algorithms are crucial. Examples are safety-critical areas in robotics, medical imaging, autonomous systems - in particular autonomous driving - as well as research in general or high-quality requirements in user experience. Statistical confidence is required for reliability. To this end we derive probabilistic error bars assigned to the posterior mean $\mathbf{R}_\mathrm{m}$ below. 

From \cref{eq:surprise} we see that the posterior distribution of $\mathbf{R}$ is Gaussian, since 
only terms up to the quadratic order are contained in $\mathcal{H}(\mathbf{R}|d,s)$. The uncertaity quantification in terms of the posterior covariance is thus determined by the inverse of
\begin{align}
    \mathbf{\Sigma}^{-1}\equiv&~\frac{\partial^2\mathcal{H}(\mathbf{R}|d,s)}{\partial \mathbf{R}^2}=\sum_i\frac{\partial^2(\mathcal{H}_1)_i}{\partial \mathbf{R}^2} + \frac{\partial^2\mathcal{H}_\mathrm{prior}}{\partial \mathbf{R}^2}\\
    =&~\sum_i (s_is_i^t)\otimes \mathbf{N}_i^{-1}  + (\mathbf{V}^t\otimes\mathbf{U})^{-1}.
    \label{eq:u}
\end{align}
Note, that for $\mathbf{N}_i=0$ the covariance $\mathbf{\Sigma}=0$, i.e. zero uncertainty in the reconstruction. For elements $\mathbf{N}_i$ larger than zero we see from equation (\ref{eq:u}) that the total uncertainty, given by $\mathbf{\Sigma}$, is decreasing with the number of points $\mathrm{n}$. For the simplified observation model, \cref{eq:map-full}, the covariance becomes
\begin{align}
    \mathbf{\Sigma}^{-1}_{a)}=\sigma_n^{-2}\sum_i (s_is_i^t)\otimes \mathds{1}^{k\times k}  + \sigma_v^{-2}\sigma_u^{-2}\mathds{1}^{k^2\times k^2}.
    \label{eq:u-simple}
\end{align}

Given mean and covariance of $\mathbf{R}$ we are now able to provide the exact posterior probability distribution in vectorized form,
\begin{equation}
    \mathcal{P}\left(\mathbf{R}|d,s)=\mathcal{G}(\mathrm{vec}(\mathbf{R})-\mathrm{vec}(\mathbf{R}_\mathrm{m}),\mathbf{\Sigma}\right).
\end{equation}
The $1\sigma-$ uncertainty interval $\mathcal{U}_{1\sigma}$ with respect to the posterior mean $\mathbf{R}_\mathrm{m}$ is given by 
\begin{equation}
    \mathcal{U}_{1\sigma} = \left[\mathrm{vec}(\mathbf{R}_\mathrm{m})-\sqrt{\mathrm{diag}(\mathbf{\Sigma})}  ~,~\mathrm{vec}(\mathbf{R}_\mathrm{m})+\sqrt{\mathrm{diag}(\mathbf{\Sigma})}\right],
    \label{eq:uncertain}
\end{equation}
where $\mathrm{diag(\mathbf{\Sigma})}$ denotes a vector, given by the diagonal of the matrix $\mathbf{\Sigma}$.  

\subsection{Demonstration on a simple example}
\label{subsec:toyexample2d}
\begin{figure}[ht]
    \centering
     \includegraphics[width=1\linewidth]{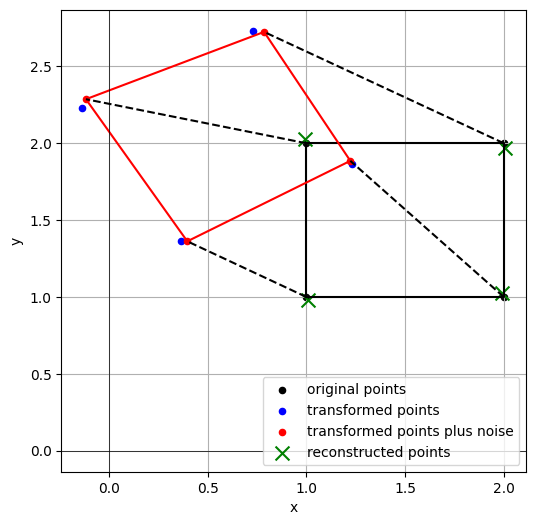}
    \caption{Original black key points $s_i$ within image $\mathcal{S}$ are observable as red key points $d_i$ in image $\mathcal{D}$ under a different perspective. Key point detection was noisy ($\sigma^2_n$), illustrated when comparing it to the noiseless (blue) points. The probabilistically inferred inverse transformation $\mathbf{R}^{-1}_\mathrm{m}$ applied to red points $d_i$ yields the reconstructed (green) crosses and proofs the concept. }
    \label{fig:poc}
\end{figure}
We provide a first proof of the derived concept with a simple example in $2$ dimensions and without using homogeneous coordinates. An experiment setup with projective transformations and homogenous coordinates is described in \cref{subsec:toyhom}. Assume we identified four key points in the images $\mathcal{S}$ (original) and $\mathcal{D}$ (other perspective). We do not know the transformation between the keypoints and have no prior knowledge about the transformation, i.e. $\sigma^{-2}_u=\sigma^{-2}_v =0$. We further assume to know how precise the keypoint detection algorithm works, i.e. the noise covariance $\sigma^2_n>0$ is known. For the sake of convenience we choose a rotation matrix, parametrized by rotation angle $\Theta$. 

For a simple experiment, we choose $\Theta=\pi/6$, $\sigma^2_n=10^{-2}$ to generate mock data,
\begin{equation}
    d^\mathrm{mock}=\mathbf{R}^\mathrm{mock}s+n=
    \begin{bmatrix}
        0.87 & -0.5\\
        0.5         & 0.87
    \end{bmatrix}s
    +n.
\end{equation}
Reconstruction of transformation matrix $\mathbf{R}_\mathrm{m}$, given mock data points $d^\mathrm{mock}$ and $s$, by applying equation (\ref{eq:simple}) and equations (\ref{eq:u-simple},\ref{eq:uncertain}) for uncertainty quantification, yields
\begin{equation}
    \mathbf{R}_\mathrm{m}=
    \begin{bmatrix}
        0.86 & -0.48\\
        0.49 & 0.89
    \end{bmatrix}
    \pm 
    \begin{bmatrix}
        0.02 & 0.02\\
        0.02 & 0.02\\
    \end{bmatrix}.
\end{equation}

\cref{fig:poc} visualizes the results of this experiment and serves as first \gls*{poc} for optimally inferring the transformation from noisy data.

\section{Image and perspective noise}
\label{sec:noisemodel}
Uncertainty in homographic projections can arise from different sources and at different stages of the imaging process. Conventional feature-localization uncertainty is introduced in the image plane and is therefore modeled as additive noise after perspective normalization (details below). Before that, deviations of the underlying imaging geometry already act on the projective mapping itself. Examples include uncertainty in camera intrinsics or extrinsics, uncertainty in a previously estimated homography, temporal misalignment between cameras observing a dynamic scene, and small geometric deviations of the observed surface from the assumed planar model. Thus, additive homogeneous noise is not introduced as an alternative representation of pixel noise, but as a local stochastic model of uncertainty in the geometry or in the physical process underlying the projection.

The homogeneous measurement model is given by
\begin{align}
    d_\mathrm{H}
    =
    \mathbf{R}s_\mathrm{H}
    +
    n_\mathrm{H} ~~,
    \label{eq:homogeneousnoise}
\end{align}
where $s_\mathrm{H}$ and $d_\mathrm{H}$ denote a single matched pair of homogeneous source and target keypoints, respectively, and
$
    n_\mathrm{H} = \begin{bmatrix}n_x & n_y & n_w\end{bmatrix}^t \sim \mathcal{G}\left(0,\mathbf{N}_\mathrm{H}\right)$, $\mathbf{N_H} \in \mathbb{R}^{(3\times3)}
$
denotes the noise in homogeneous coordinates. Note that for simplicity, the point index $i$ is omitted throughout this section unless explicitly required. All derived relations apply individually to each pair of matched points.

The homogeneous projection $\mathbf{R}s_\mathrm{H}$ contains the projective component
$
    w
    :=
    (\mathbf{R}s_\mathrm{H})_3 =
    r_{31}s_x
    +
    r_{32}s_y
    +
    r_{33}
$, 
which is used for perspective normalization.
Note that $r_{33}=1$ in all cases due to normalization of the homography matrix. 

Projection of \cref{eq:homogeneousnoise} onto the two-dimensional Euclidean image plane yields
\begin{align}
    d_\mathrm{P}
    =
    \mathrm{P}(d_\mathrm{H})
    =
    \frac{
        (\mathbf{R}s_\mathrm{H})_{[1:2]}
        +
        (n_\mathrm{H})_{[1:2]}
    }{
        w+n_w
    } ~~,
    \label{eq:homogenousenoise2}
\end{align}
and introduces non-linearity into the measurement model, where subscript $\mathrm{P}$ denotes projected quantities and $\mathrm{P}(\cdot)$ is the projection operator.
Pixel noise introduced directly in the two-dimensional Euclidean image space is a common special case in image-based \gls*{cv} problems. While the homogeneous measurement model, \cref{eq:homogeneousnoise}, with $n_w=0$ remains applicable in close-to-affine scenarios ($w \approx 1$) and low noise cases, $n \ll 1$, we have to account for non-linearities otherwise. 

Keypoint observations are commonly modeled by additive Gaussian noise \cite{hartleyMultipleViewGeometry2004, zhangFlexibleNewTechnique2000},
\begin{align}
    d_\mathrm{P}
    =
    \mathrm{P}\left(
        \mathbf{R}\tilde{s}_\mathrm{P}
    \right)
    +
    n_\mathrm{P},
    \qquad
    n_\mathrm{P}
    \sim
    \mathcal{G}(0,\mathbf{N}_\mathrm{P}),
    \label{eq:perspectivenormalization}
\end{align}
where $s_\mathrm{P}$ and $d_\mathrm{P}$ denote a single matched pair of source and target image keypoints, respectively, $\tilde{\cdot}$ denotes the normalized homogeneous representation of an image point.

To model pixel noise we set $n_w=0$ in \cref{eq:homogenousenoise2}.
Comparison with \cref{eq:perspectivenormalization} gives
$
    (n_\mathrm{H})_{[1:2]}
    =
    w n_\mathrm{P}
$ and therefore

\begin{align}
    \left(\mathbf{N_\mathrm{H}}\right)_{\left[1:2, 1:2\right]}
    =
    w^2\mathbf{N}_\mathrm{P} ~~.
    \label{eq:pixeltohomogeneouscovariance}
\end{align}
Thus, Gaussian measurement noise specified in image coordinates can be represented in homogeneous coordinates by scaling its covariance with the squared projective component $w$. Considering \cref{eq:pixeltohomogeneouscovariance} and after simple rearranging, \cref{eq:perspectivenormalization} becomes

\begin{align}
\tilde{d}_\mathrm{P} = \frac{1}{w} \mathbf{R}\tilde{s}_\mathrm{P}+\tilde{n}_\mathrm{P} ~~.
\label{eq:projectedemodel}
\end{align}

Since $w$ depends on $\mathbf{R}$ and each signal point $s_i$ itself, \cref{eq:projectedemodel} is still a nonlinear inference problem that might be solved by methods of variational inference. To still use the linear model derived in \cref{sec:methodology}, we propose the following approach: We initially assume that $\mathbf{R}$ and $w$ are independent and alternately infer $\mathbf{R}$ using \cref{eq:wienerf} for a fixed $1/w$, and $1/w$ for a fixed $\mathbf{R}$. During this process, the mean of $1/w$ is also estimated by a Wiener filter in each iteration step,

\begin{align}
\left(\frac{1}{w_i}\right)_m &= \left[{W}^{-1}_{\mathrm{inv,i}}+(x^i)^t\mathbf{N}^{-1}_\mathrm{P}(x^i)\right]^{-1} \\
&\times\left((x^i)^t\mathbf{N}^{-1}_\mathrm{P}\tilde{d}^i_\mathrm{P} + {W}_{\mathrm{inv,i}}^{-1} \overline{\left(\frac{1}{w_i}\right)}\right) ~~\forall i ~~,
\label{eq:wienerw}
\end{align}
where $W_{\mathrm{inv},i}$ denotes the prior variance of $1/w_i$, $\overline{(1/w_i)}$ its prior mean, and $x^i \equiv \mathbf{R}\tilde{s}_\mathrm{P}^i$.
This process repeats until convergence, e.g., when the changes in $\mathbf{R}_\mathrm{m}$ between iterations fall below a certain threshold. By initializing the approach with a good choice for $\mathbf{R}$ (e.g. the \gls*{dlt} solution), convergence is accelerated.

\subsection*{Non-Gaussian noise}
Although the derivation in \cref{sec:methodology} uses a Gaussian noise model, the results are valid for other noise models. Let $\mathcal{P}_n$ be a general noise distribution. Then the likelihood is given by 
\begin{align}\mathcal{P}(d|\mathbf{R},s) = \mathcal{P}_n(d-\mathbf{\hat{R}}s)~~.\end{align}
To keep the approach analytically, but without loss of generality for non-Gaussian noise, one might use a conjugate prior with regard to the noise probability distribution 
$\mathcal{P}_\mathrm{conjugate}(\mathbf{R})$
to include prior knowledge of $\mathbf{R}$. Hence, the posterior information reads 
\begin{align}
\mathcal{H}(\mathbf{R}\mid d,s)
=
-\ln \mathcal{P}_n(d-\mathbf{\hat{R}}s) \label{eq:generalinformation1}
-\ln \mathcal{P}_{\mathrm{conjugate}}(\mathbf{R})
+ \mathcal{H}_0.
\end{align}

Following \crefrange{subsec:bayesianofthehomographic}{subsec:uq} one might derive analytical expressions for the posterior mean or maximum, depending on the specific noise/prior distribution used.

\section{Experimental Setup}
\subsection{Synthetic Data}
\label{subsec:toyhom}
This proof of concept is an extension of \cref{subsec:toyexample2d} in homogeneous coordinates and for projective transformations. Again, all parameters ($\mathbf{R}^\mathrm{mock}$, $s_\mathrm{H}$ and $d_\mathrm{H}^\mathrm{mock}$)  are known. 
The noise is modeled according to \cref{sec:noisemodel}. Noise levels are given in the units of the synthetic coordinate system.
For numerical stability, Hartley normalization \cite{hartleynorm1997,hartleyMultipleViewGeometry2004} is applied to the noisy keypoints before homography estimation.

\begin{figure*}
  \centering
  \begin{subfigure}{0.49\linewidth}
        \centering
        \includegraphics[width=\linewidth]{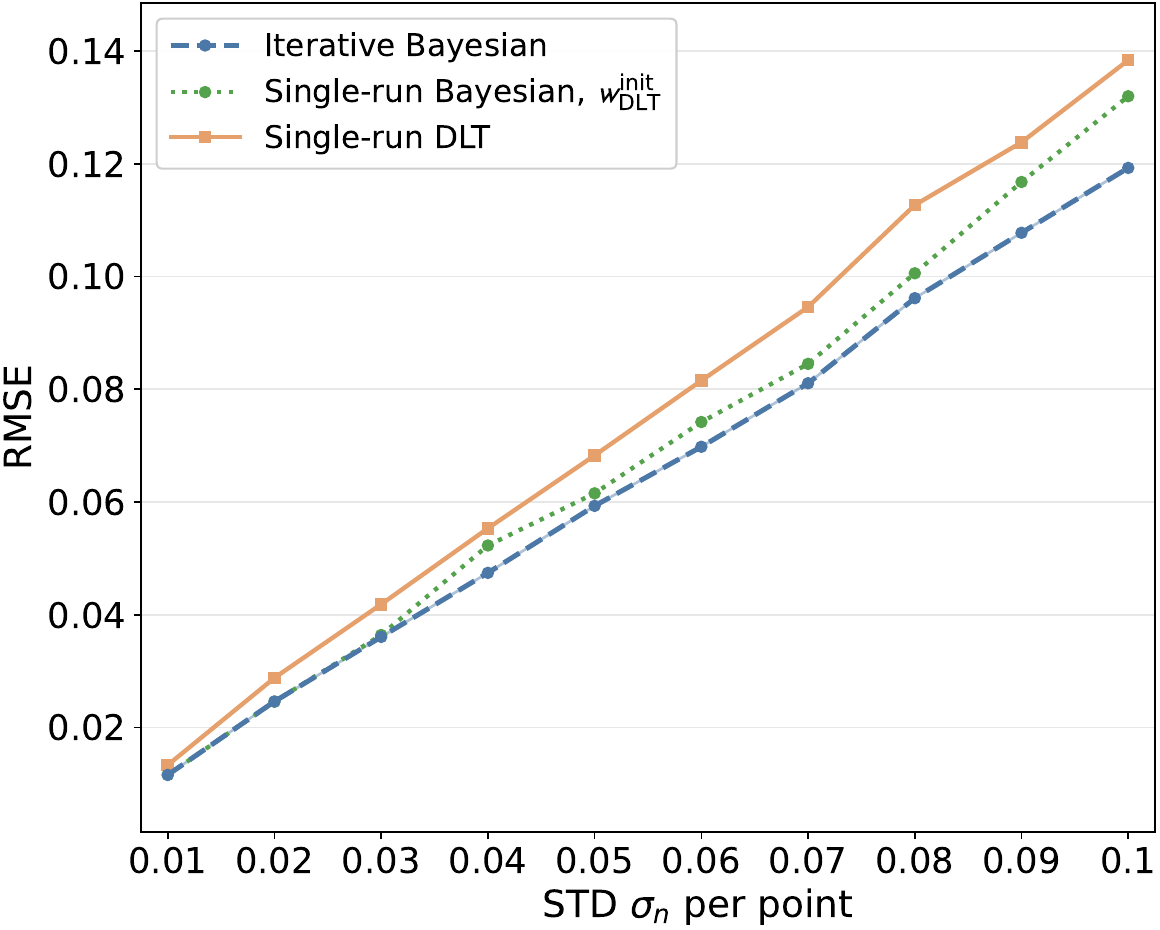}
        \caption{Mean RMSE over 100 runs.}
        \label{fig:RMSEtoyexample}
    \end{subfigure}
  \hfill
  \begin{subfigure}{0.49\linewidth}
    \includegraphics[width=1\linewidth]{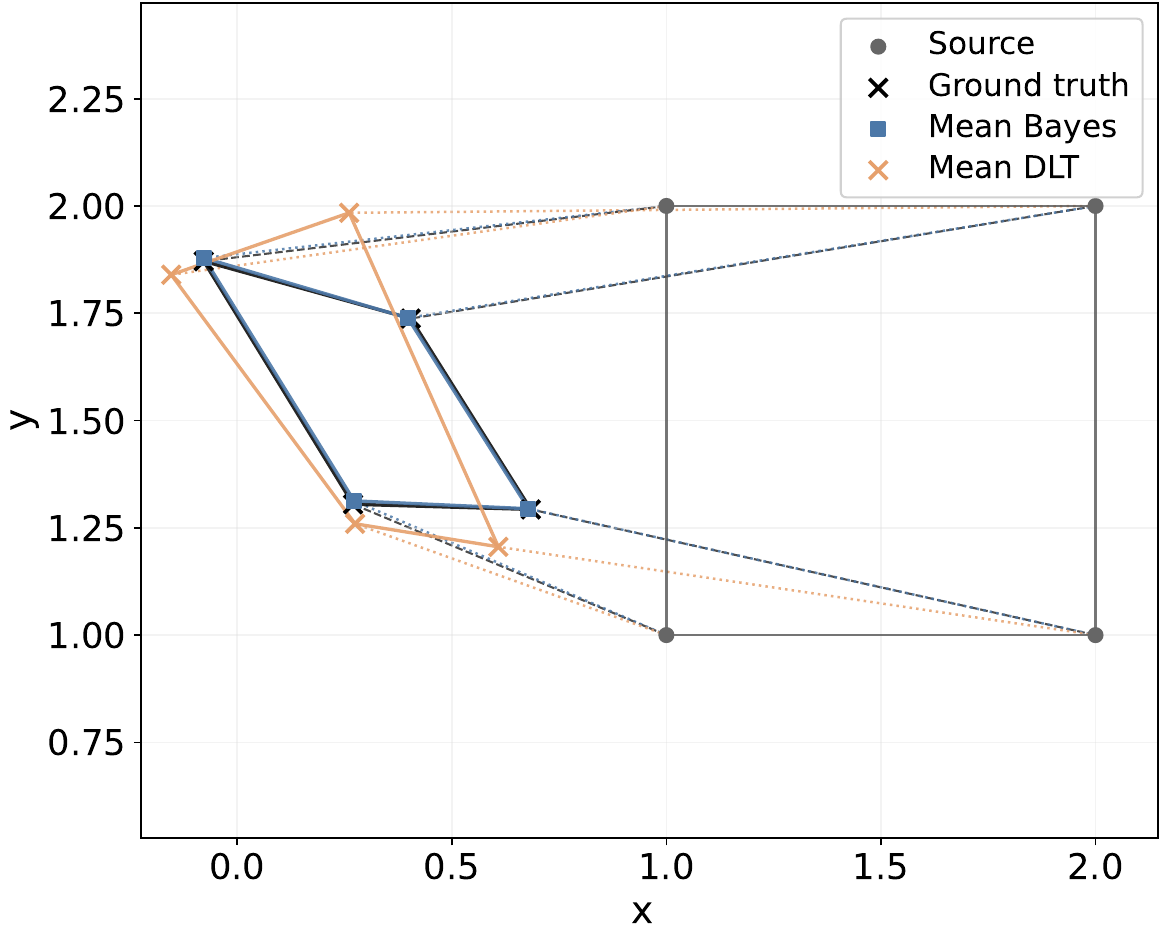}
    \caption{Mean homography matrices over 100 runs for $\sigma_\mathrm{n}=0.04$.}
    \label{fig:meanhom}
  \end{subfigure}
  \caption{Results for the experiment given a projective transformation.}
  \label{fig:toyresults}
\end{figure*}

For a given noise level, 100 independent runs are performed. In each run, a new noise vector is sampled and added to $\mathbf{R}^\mathrm{mock}s_\mathrm{H}$. A posterior mean homography matrix $\mathbf{R}_\mathrm{m}$ and its parameter uncertainty (\cref{eq:u,eq:uncertain}) are then inferred using \cref{eq:wienerf}. The iterative \gls*{bi} (\cref{eq:wienerf}, \cref{eq:wienerw}) is considered converged when between two consecutive iterations the relative change between the homography matrices and the displacement of the transformed keypoints fall below certain thresholds ($\epsilon_{\Delta\mathbf{R}}=10^{-6},~\epsilon_{\Delta\mathrm{points}} = 10^{-3}$). For comparison, \gls*{dlt} is applied to the same noisy keypoint correspondences in each run. The estimated homographies are applied to the source keypoints and compared to the noise-free target keypoints $\mathbf{R}^\mathrm{mock}s_\mathrm{H}$. The resulting \gls*{rmse} (see details in \cref{subsec:imgstitchsetup}) is calculated for each run and averaged over the 100 independent noise realizations. This procedure is repeated for several noise levels to evaluate the estimation accuracy under increasing observation noise. Results can be found in \cref{fig:toyresults} and are discussed in the subsequent section.

\subsection{Image Stitching Experiment}
\label{subsec:imgstitchsetup}
\begin{figure*}
  \centering
  \begin{subfigure}{0.49\linewidth}
        \centering
        \includegraphics[width=\linewidth]{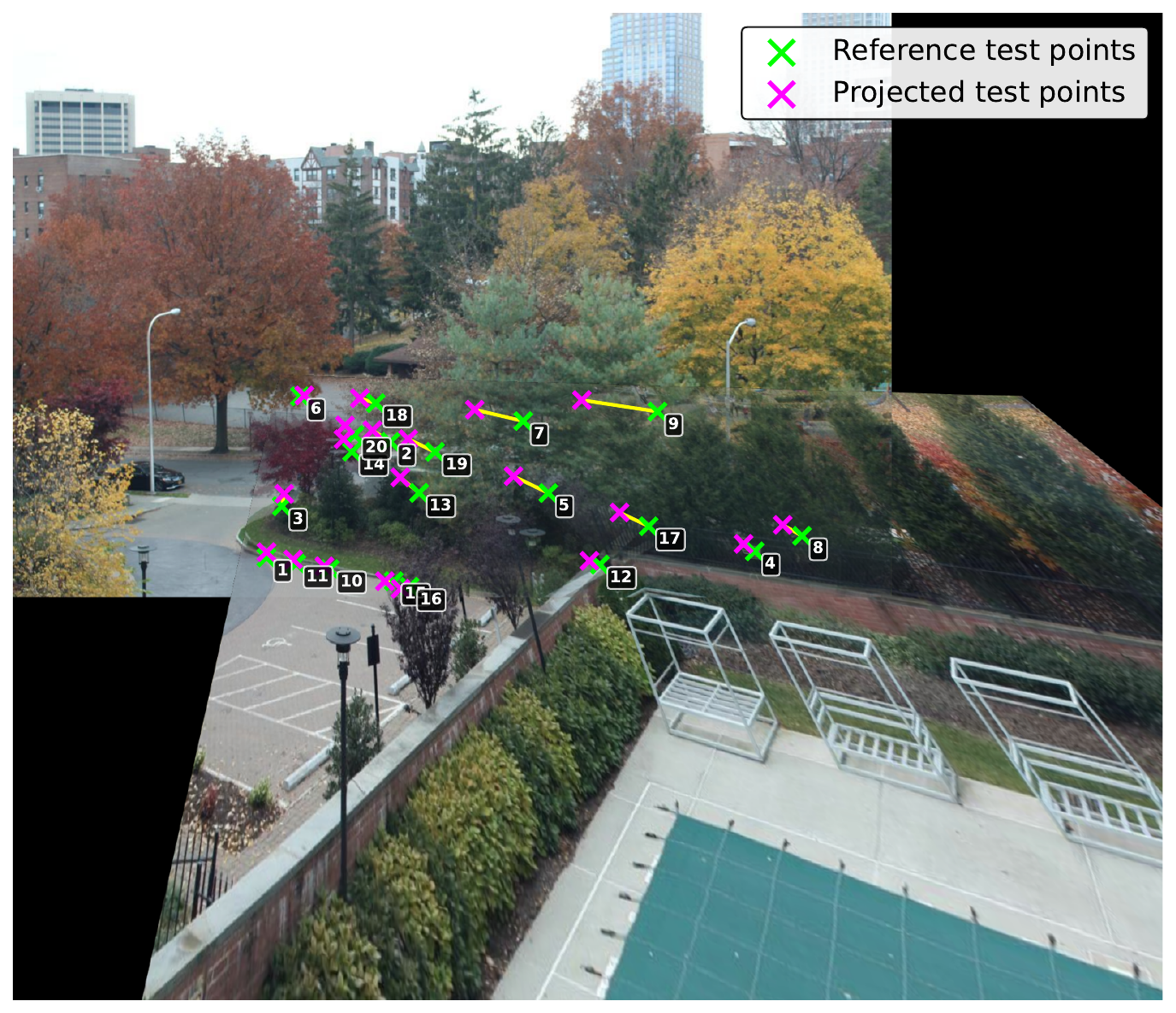}
        \caption{\gls*{dlt} estimation (\gls*{rmse}$=47.47$ px). The \gls*{dlt} solution is visibly distorted.}
        \label{fig:sky_dlttestpoints}
    \end{subfigure}
  \hfill
  \begin{subfigure}{0.49\linewidth}
    \includegraphics[width=1\linewidth]{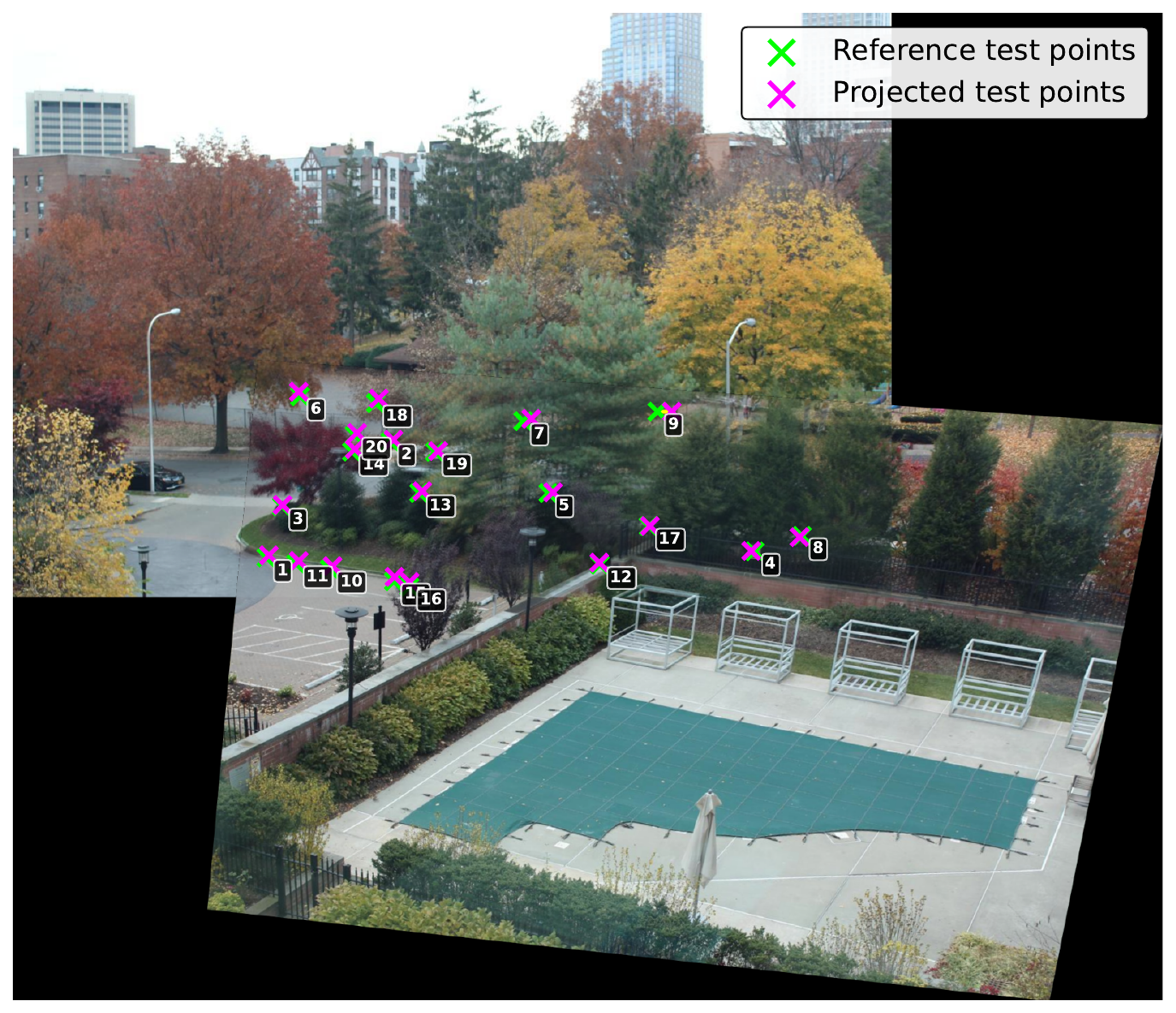}
    \caption{Iterative Bayesian approach with probabilistic estimation of ${w}$ and initialization with Bayes (\gls*{rmse}$= 8.92$ px).}
    \label{fig:sky_bayestestpoints}
  \end{subfigure}
  \label{fig:testpointpanoramas}
      \caption{Comparison of a stitched panorama with four point correspondences estimated by \gls*{dlt} (left) and our proposed iterative Bayesian approach (right) on images from the StitchBench\cite{cai2025stitchbench}/AANAP\_skyline\cite{linaanapskyline2015} dataset.}
      \label{fig:panoramas}
\end{figure*}

In a next step, we extend the \gls*{poc} to image stitching, using images from the StitchBench dataset collection \cite{cai2025stitchbench} and HPatches \cite{hpatches_2017_cvpr}. The proprosed method is still applied in a controlled environment, but now not all parameters are known, namely $\mathbf{R}$ and ${w_i}$ (see \cref{sec:noisemodel}). $s_\mathrm{P}$ and $d_\mathrm{P}$ are detected using \gls*{sift} \cite{loweDistinctiveImageFeatures2004} and are matched between the two images. In total, only four\footnote{In order to demonstrate the strength of \gls*{bi} the minimum of 4 points has been chosen.} keypoint pairs across foreground and background regions are selected and used for homography estimation. No additional outlier removal is implemented. Noise samples of different zero mean Gaussian noise are added to the homogeneous destination coordinates $\tilde{d}_\mathrm{P}$ which acts as $d_\mathrm{H}$. A second set of correspondences is selected as a test set and is not used during homography estimation. For both \gls*{dlt} and the Bayesian estimator, Hartley normalization \cite{hartleynorm1997, hartleyMultipleViewGeometry2004} is applied to $s_\mathrm{H}$, $d_\mathrm{H}$ and the estimated homographies are transformed back to the original image coordinate system for evaluation.
The probabilistic estimation of $(1/w)_\mathrm{m}$ and $\mathbf{R}_\mathrm{m}$ is done as described in \cref{sec:noisemodel}.  
$\mathbf{R}^{\mathrm{init}}$ is initialized using either \gls*{dlt} or a single run using the Bayesian estimator. For the estimation of $\mathbf{R}_\mathrm{Bayes}^\mathrm{init}$ a zero mean prior is assumed, because of a lack of further information on the true homography matrix. Regardless of the initialization method, $\mathbf{R}^\mathrm{init}$ is used for the initialization of $\hat{w}$ and as the prior mean for $\mathbf{R}$ during inference.
Performance is evaluated using the two-dimensional \gls*{rmse} between the original destination keypoints and the corresponding source keypoints of the test data set transformed by the estimated homographies. The Bayesian estimator additionally reports the posterior $1\sigma$ standard deviations of the inferred homography parameters as parameter uncertainty. Convergence of the iterative estimator is controlled according to the criteria introduced in \cref{subsec:toyhom}. See \cref{fig:panoramas} and Appendix \ref{app:imagestitching} for representative examples and \cref{sec:results} for further discussion.
\section{Results}
\label{sec:results}
\subsection{Synthetic Data}
\label{subsec:toyresult}
The experiment was performed for the complete range of considered noise levels $\mathbf{N}_\mathrm{P} = \mathbf{diag}(\sigma_{n}^2), \sigma_{n} \in \{0.01,\,0.02,\ldots,\,0.10\}$ and the ground truth matrices,  
\begin{equation}
\begin{aligned}
\mathbf{R}^\mathrm{mock}_\mathrm{affine} =
&\begin{bmatrix}
0.86 & -0.50 & 0.02\\
0.50 & 0.86 & 0.50\\
0.00 & 0.00 & 1.00
\end{bmatrix}~~, \\ \mathbf{R}^\mathrm{mock}_\mathrm{projective} =
&\begin{bmatrix}
0.86 & -0.50 & 0.02\\
0.50 & 0.86 & 0.50\\
0.40 & 0.03 & 1.00
\end{bmatrix}~~,
\end{aligned}
\label{eq:truematrices}
\end{equation}
 with results being aggregated over 100 runs per noise level. Note that as discussed in \cref{sec:noisemodel}, $n_w=0$, but $(\mathbf{N}_\mathrm{H})_{33}\approx0$ to avoid matrix singularity. To reflect only little knowledge about $\mathbf{R}$, high values for the prior parameters are chosen: $\mathbf{U_\mathrm{affine}}=\mathbf{diag}(10,10,10^{-4})$, $\mathbf{V_\mathrm{affine}} =\mathbf{diag}(1,1,2.5)$ and $\mathbf{U_\mathrm{projective}}=\mathbf{diag}(10,10,2.5)$, $\mathbf{V_\mathrm{projective}} =\mathbf{diag}(1.0)$. We modify this assumption in the image stitching proof of concept as described in \cref{subsec:imgstitchsetup}. In the projective case, the third row of $\mathbf{R}$ should not be pushed towards an affine solution, which is why $\mathbf{U}_{33}, \mathbf{V}_{33}$ differ in the affine and projective cases.

\cref{fig:toyresults} shows results for the projective case.
The graphs in \cref{fig:RMSEtoyexample} show the mean \gls*{rmse} for the different estimation methods per noise level.
\cref{fig:RMSEtoyexample} depicts a standard \gls*{dlt} solution (orange), a single-run \gls*{bi} estimation, where the \gls*{dlt} solution has been used for initialization of $w$ (\cref{sec:noisemodel}) (green) and the proposed iterative \gls*{bi} (blue) method. For the iterative \gls*{bi} method ${w}$ has been initialized separately with $1$ and $w_\mathrm{DLT}^\mathrm{init}$. Both initializations lead to non-distinguishable overlapping curves, which is why only a single iterative Bayesian curve is shown.
Accuracy improves with iterative optimization, whereas the single-run \gls*{bi} method performs between the iterative approach and standard \gls*{dlt} while additionally providing uncertainty quantification. Although all three methods perform better in lower noise levels, the difference between \gls*{dlt} and \gls*{bi} increases, with our proposed method achieving lower \gls*{rmse}.
The ground truth transformation (black) in \cref{fig:meanhom} is accurately matched by the mean homography inferred with our proposed method (blue). The orange \gls*{dlt} solution under noise differs considerably from the real transformation. This is expected because our proposed method shows lower spread across the individual homography parameters and \gls*{dlt} shows extreme cases in outliers for each matrix parameter, which leads to the distorted transformations. Comparing the mean estimates over 100 runs, the iterative Bayesian method is considerably closer to the ground-truth homography than the \gls*{dlt} estimate.
See Appendix \ref{app:synth_projtrans} for the numerical parameters and Appendix \ref{app:synth_afftrans} for the affine experiment results. 
This \gls*{poc} in a controlled environment demonstrates that the theory behind the proposed approach is solid even for projective transformations. 

\subsection{Image Stitching Example}
\label{subsec:imgstitchresult}
The image sets used are AANAP/skyline \cite{linaanapskyline2015}, NISwGSP-06\_PalazzoPubblico \cite{Chen:2016:NIS}, OBJ-GSP-03\_river, Aerial/Restaurant, included in the StitchBench dataset collection \cite{cai2025stitchbench}, and HPatches/v\_graffiti \cite{hpatches_2017_cvpr}. For AANAP/skyline, Gaussian noise with $\sigma_n = 10.0\,\mathrm{px}$ is added to $d_\mathrm{H}$. The \gls*{dlt} solution (\cref{fig:sky_dlttestpoints}) shows visible distortion with an \gls*{rmse} of $47.42\,\mathrm{px}$ and a runtime of $1\,\mathrm{ms}$. With empirically selected parameters, initialization using $\hat{w}^\mathrm{init}_\mathrm{Bayes}$, and a prior of $(\mathbf{V}\otimes\mathbf{U})=\mathbf{diag}(0.1)$, the iterative Bayesian approach converges after $184$ iterations in $2.18\,\mathrm{ms}$ and reduces the \gls*{rmse} to $8.92\,\mathrm{px}$ (\cref{fig:sky_bayestestpoints}). The corresponding homography matrices are provided in Appendix \ref{app:skyline_results}, along with stitching results other than those shown in \cref{fig:panoramas} (Appendix \ref{app:vgraffiti}--\ref{app:palazzo}).

For v\_graffiti, Gaussian noise with $\sigma_n=5.0$ px is used and the iterative approach is initialized with \gls*{dlt} without a prior. While \gls*{dlt} requires $1.43\,\mathrm{ms}$ and reaches an \gls*{rmse} of $16.95\,\mathrm{px}$, the iterative approach requires $1507$ iterations and $1679\,\mathrm{ms}$ but improves the initial solution to $14.11\,\mathrm{px}$. Thus, despite initialization with \gls*{dlt}, the iterative estimation further optimizes the homography based on the introduced projective information. Detailed matrices and the resulting panorama are provided in Appendix \ref{app:vgraffiti}.

With this \gls*{poc}, we have demonstrated, that the iterative Bayesian approach with pixel noise is able to outperform the \gls*{dlt} method, even though we introduced the initial independence of $w$ and $\mathbf{R}$. If the proposed method is efficiently initialized, it can derive better results as the initialization method. In scenarios where the iterative method is not improving from the initial \gls*{dlt} guess, our approach still adds uncertainty quantification to the mean estimator.

\section{Conclusion}
This work presented a Bayesian formulation for homography estimation that incorporates measurement uncertainty and prior knowledge and provides posterior estimates of the homography parameters and their uncertainty. A closed-form solution has been derived for the posterior mean in the linear homogeneous-noise model and has been extended by an iterative approach to account for the nonlinearities introduced by perspective normalization in the pixel-noise case. Proof-of-concept experiments on synthetic data and real image correspondences demonstrate the applicability of the proposed formulations under their respective assumptions. It demonstrates the feasibility of Bayesian homography estimation with explicit incorporation of measurement and parameter uncertainty. 

The further development of the iterative Bayesian \gls*{poc}-method for the nonlinear pixel-noise case to some direct Bayesian method, followed by an intensive benchmark comparison using this direct method is left for future work. Additionally, research on the application of the presented Bayesian method to uncertainty in the geometry or in the physical process underlying the projection is planned.

\section*{Acknowledgements}
This research was funded by the Free State of Bavaria, Germany, through the Hightech Agenda Bayern, and supported by the Technical University of Applied Sciences Augsburg and its Technology Transfer Center (TTZ) Data Science and Autonomous Systems Landsberg am Lech. The authors would also like to thank Adrian Schlosser for his valuable feedback on the manuscript.

{
    \small
    \bibliographystyle{ieeenat_fullname}
    \bibliography{main}
}
\appendix

\section{Synthetic results}
\label{app:synthetic_results}

\subsection{Affine transformation}
\label{app:synth_afftrans}

\begin{figure}[!ht]
 \centering
 \includegraphics[width=1\linewidth]{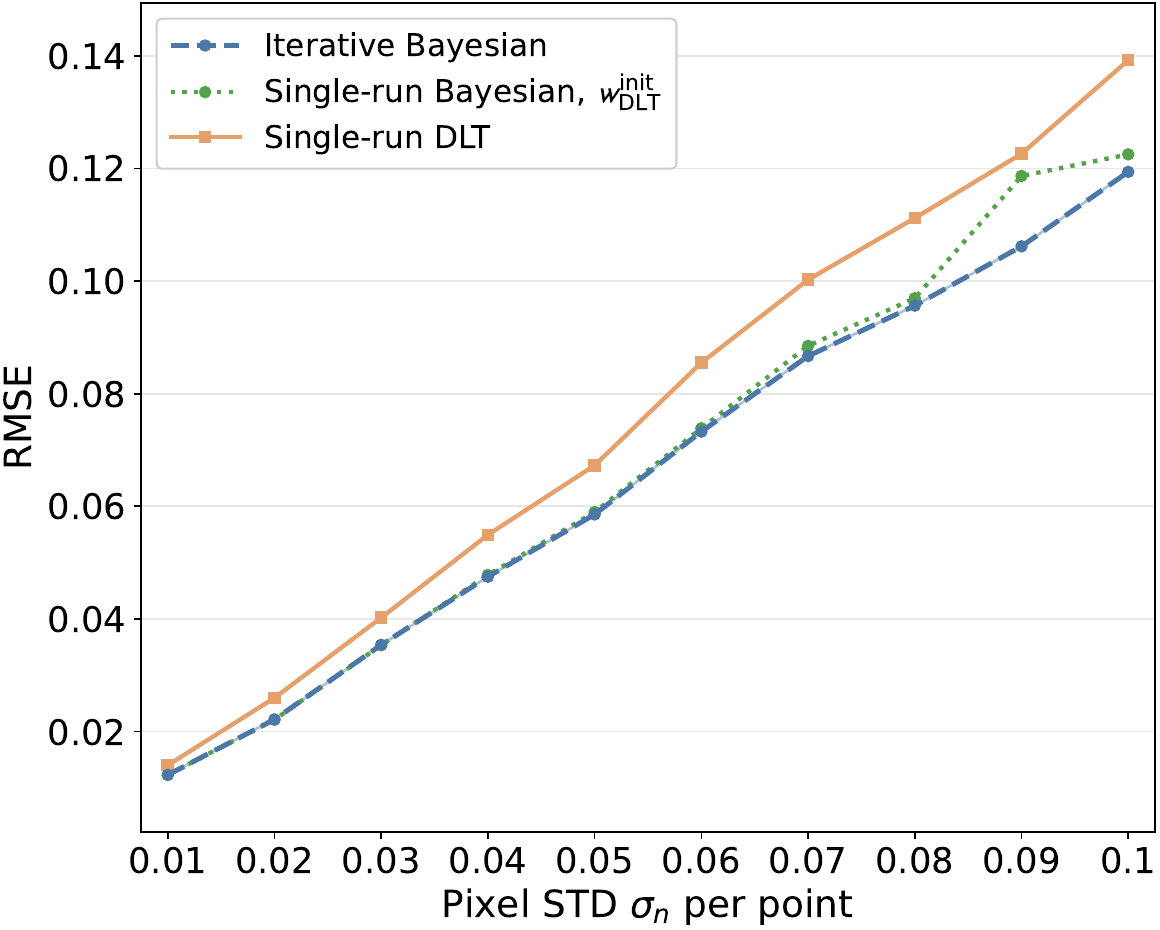}
        \caption{Affine scenario: Mean RMSE over 100 runs. Iterative Bayesian (Blue) performs best, while Single run Bayesian ($w^\mathrm{DLT}_\mathrm{init}$) performs similarly, unlike in the projective case.} 
        \label{appfig:RMSEtoyexample}
\end{figure}

The inferred Bayesian parameters $\mathbf{R}_\mathrm{m}^\mathrm{Bayes}$\footnote{\label{fn:paramnorm}Only the normalized homography matrix is reported, even though the $1\sigma$ values are derived for the non-normalized homography matrix. If not declared otherwise, the inferred homography is near $(\mathbf{R}_{33})=1$.} and the DLT solution $\mathbf{R}^\mathrm{DLT}$ for the affine case at $\sigma_n=0.04$ are
\begin{align}
\mathbf{R}^\mathrm{DLT} = 
&\begin{bmatrix}
    0.93 & -0.52 & -0.02 \\
    0.58 &0.98 &0.38 \\
    0.02 &0.03 &1.00
\end{bmatrix}~~,\\
\mathbf{R}^\mathrm{Bayes}_\mathrm{m} =
&\begin{bmatrix}
          0.87 &
          -0.50&
          0.02\\
          0.50 &
          0.87 &
          0.50\\
          0.00 &
          0.00 &
          1.00
\end{bmatrix} \\
\pm &\begin{bmatrix}
          0.04 &
          0.04 &
          0.09 \\
          0.04 &
          0.04 &
          0.09\\
          0.00&
          0.00&
          0.00
\end{bmatrix}.
\end{align}
In contrast to the projective case, the single-run BI follows the iterative solution more closely, as the affine transformation does not introduce a projective scale $w$. Iterative optimization consistently achieves the lowest RMSE. At $\sigma_n=0.04$, the Bayesian estimate closely recovers the ground-truth affine transformation, while the DLT estimate shows larger parameter deviations.
Isotropic noise yields identical uncertainties in the first two rows, while $(\mathbf{N}_\mathrm{H})_{33} \ll 1$ introduces small near-zero uncertainties in the third row despite $n_w=0$.

\subsection{Projective transformation}
\label{app:synth_projtrans}
The inferred Bayesian parameters $\mathbf{R}_\mathrm{m}^\mathrm{Bayes}$ and the DLT $\mathbf{R}^\mathrm{DLT}$ solution of the projective case at $\sigma_n = 0.04$ are
\begin{align}
\mathbf{R}^\mathrm{DLT}_\mathrm{m} = 
&\begin{bmatrix}
    0.14 & -0.24 & 0.28 \\
    -0.18 &-0.05 &1.04 \\
    -0.12 &-0.23 &1.00
\end{bmatrix}~~,\\
\mathbf{R}^\mathrm{Bayes}_\mathrm{m} =
&\begin{bmatrix}
          0.85 &
          -0.50&
          0.04\\
          0.49 &
          0.87 &
          0.52\\
          0.40 &
          0.03 &
          1.00
\end{bmatrix} \\
\pm &\begin{bmatrix}
          0.07 &
          0.06 &
          0.14 \\
          0.07 &
          0.06 &
          0.14 \\
          0.00&
          0.00&
          0.00
\end{bmatrix}.
\end{align}
The uncertainty structure follows the same noise assumptions as in the affine case.

\section{Image stitching results}
\label{app:imagestitching}

\subsection{AANAP/skyline}
\label{app:skyline_results}
The numerical transformation parameters for the AANAP\_skyline \cite{linaanapskyline2015} images from the StitchBench \cite{cai2025stitchbench} dataset collection are
\begin{align}
\mathbf{R}^\mathrm{DLT} = &\begin{bmatrix}
    0.54 & -0.50 & 464.63 \\ 
    -0.09 & 0.17 & 631.97 \\
    0.00 & 0.00 & 1.00
    \end{bmatrix}~~, \\
\mathbf{R}_\mathrm{m}^\mathrm{Bayes} = &\begin{bmatrix}
        0.96 & -0.06 & 426.58 \\
        0.05 & 1.05 & 605.47 \\
        0.00 & 0.00 & 1.00  
\end{bmatrix} \\ \pm &\begin{bmatrix}
        0.01 & 0.03 & 0.32 \\
        0.01 & 0.03 & 0.32 \\
        0.01 & 0.01 & 0.32
\end{bmatrix}.
\end{align}
For the resulting panoramas see Fig.~\ref{fig:panoramas} in the main paper. The low uncertainty is a result of the chosen prior.

\subsection{HPatches/v\_graffiti}
\label{app:vgraffiti}
\begin{figure*}
  \centering
   \begin{subfigure}[t]{0.49\linewidth}
        \centering
        \includegraphics[width=\linewidth]{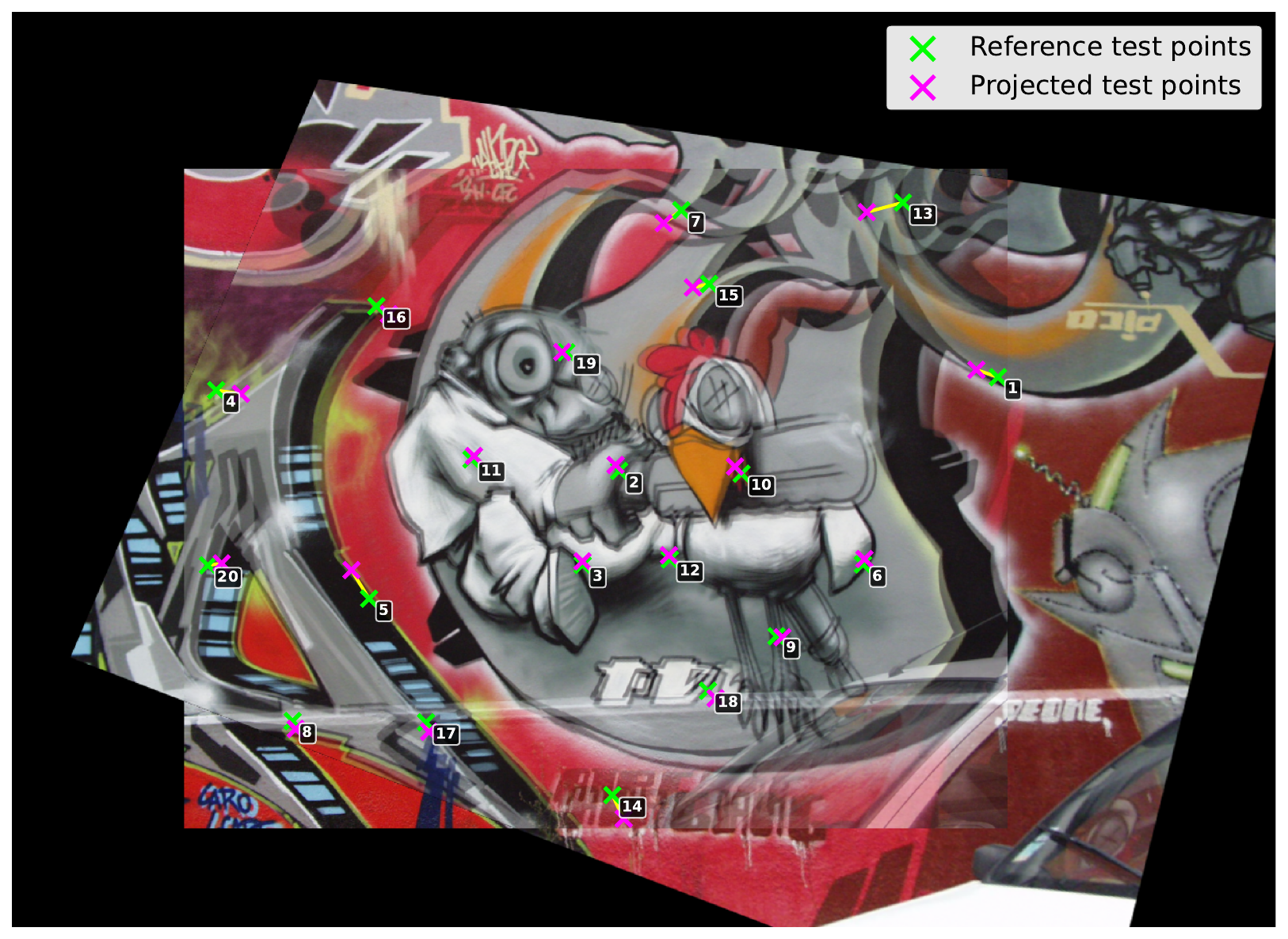}
        \caption{DLT estimation (RMSE$=16.95$ px).}
    \end{subfigure}
     \begin{subfigure}[t]{0.49\linewidth}
        \centering
        \includegraphics[width=\linewidth]{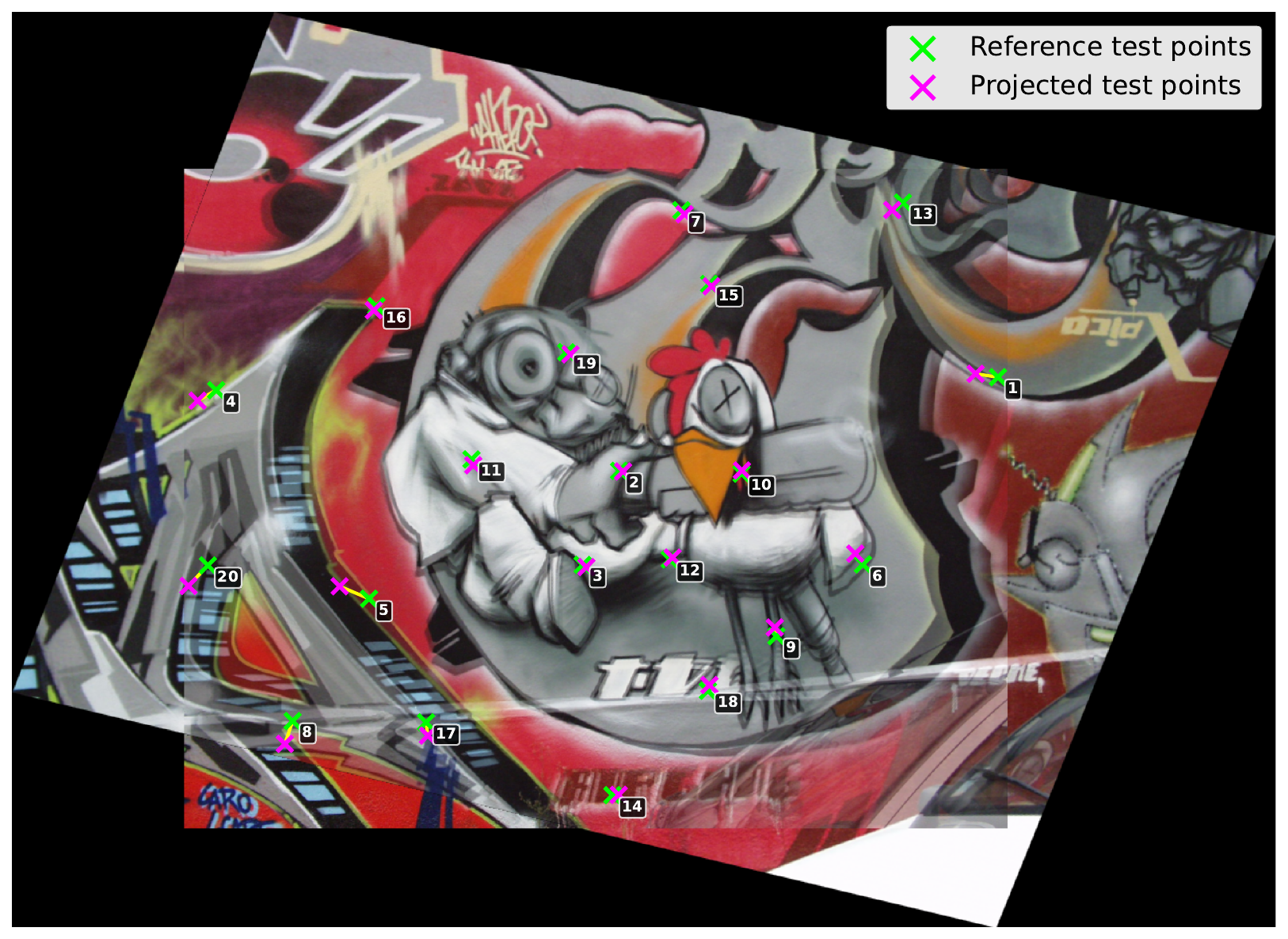}
        \caption{Iterative Bayesian approach (RMSE $=14.11\, \mathrm{px}$).}
    \end{subfigure}
\caption{Comparison of a HPatches/v\_graffiti panorama.}
 \label{appfig:v_graffiti}
\end{figure*}

The numerical transformation parameters for the HPatches/v\_graffiti images\footnote{Images 1.ppm and 2.ppm.} \cite{hpatches_2017_cvpr} are
\begin{align}
\mathbf{R}^\mathrm{DLT} = &\begin{bmatrix}
    0.85 &-0.36 &130.70 \\
        0.16 &0.81 &-87.02 \\
        0.00 & 0.00 & 1.00
    \end{bmatrix}~~, \\
\mathbf{R}_\mathrm{m}^\mathrm{Bayes} = &\begin{bmatrix}
        1.19 &-0.40 &87.29 \\
        0.27 &1.04 &-151.80 \\
        0.00 &0.00 &1.00
\end{bmatrix} \\ \pm &\begin{bmatrix}
        0.02 &0.03 &10.79 \\
        0.02 &0.03 &10.79 \\
        0.01 &0.01 & 4.83
\end{bmatrix}.
\end{align}
The resulting panoramas are shown in \cref{appfig:v_graffiti}. Assuming isotropic noise as before, the iterative Bayesian estimate shows a lower RMSE than DLT for the considered point correspondences, with larger absolute translational uncertainties reflecting their pixel scale. Similar behavior in the following examples is not discussed further.

\subsection{OBJ-GSP-03\_river}
\begin{figure*}
  \centering
   \begin{subfigure}[t]{0.49\linewidth}
        \centering
        \includegraphics[width=\linewidth]{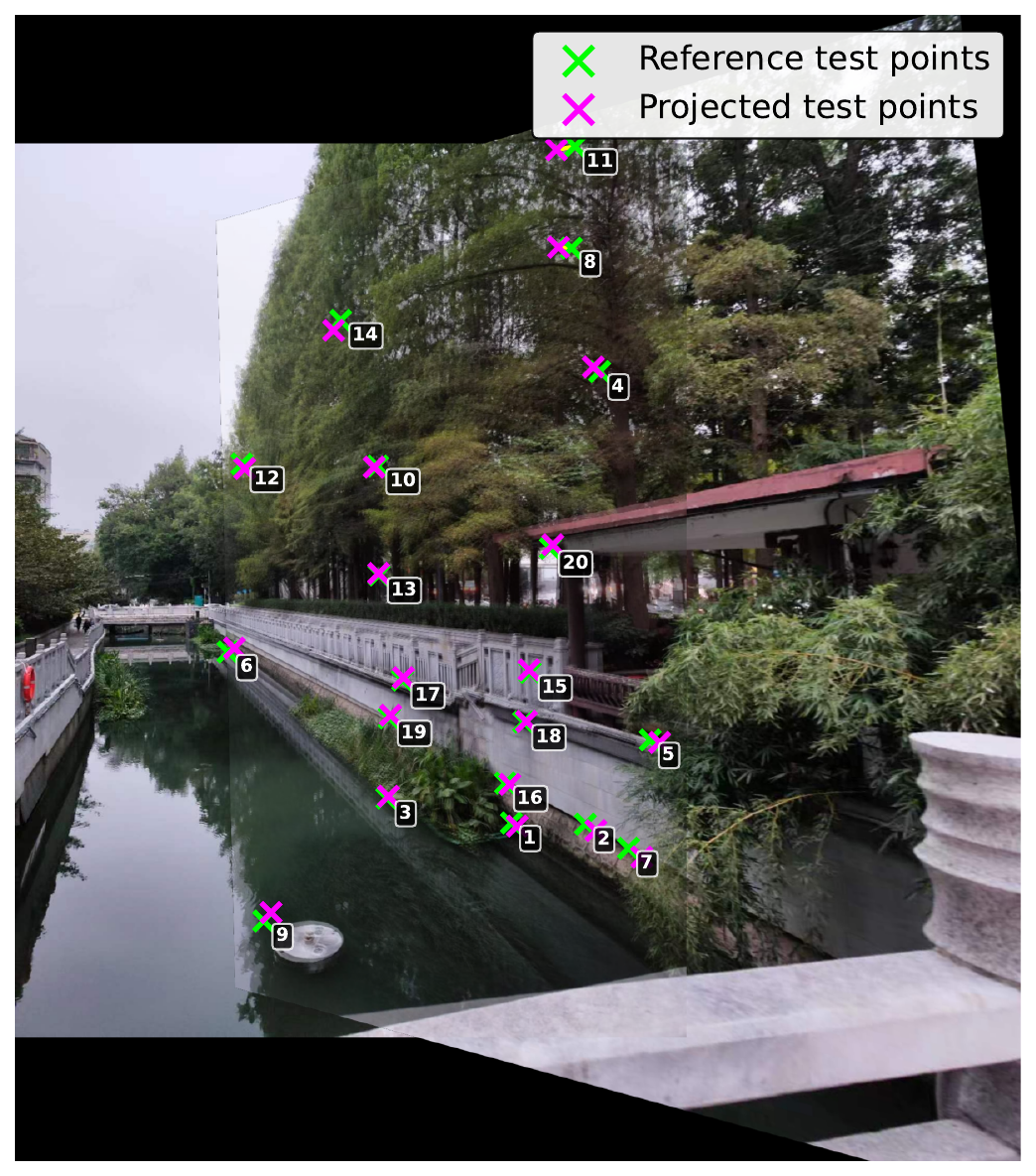}
        \caption{DLT estimation (RMSE$=17.03$ px).}
    \end{subfigure}
     \begin{subfigure}[t]{0.49\linewidth}
        \centering
        \includegraphics[width=\linewidth]{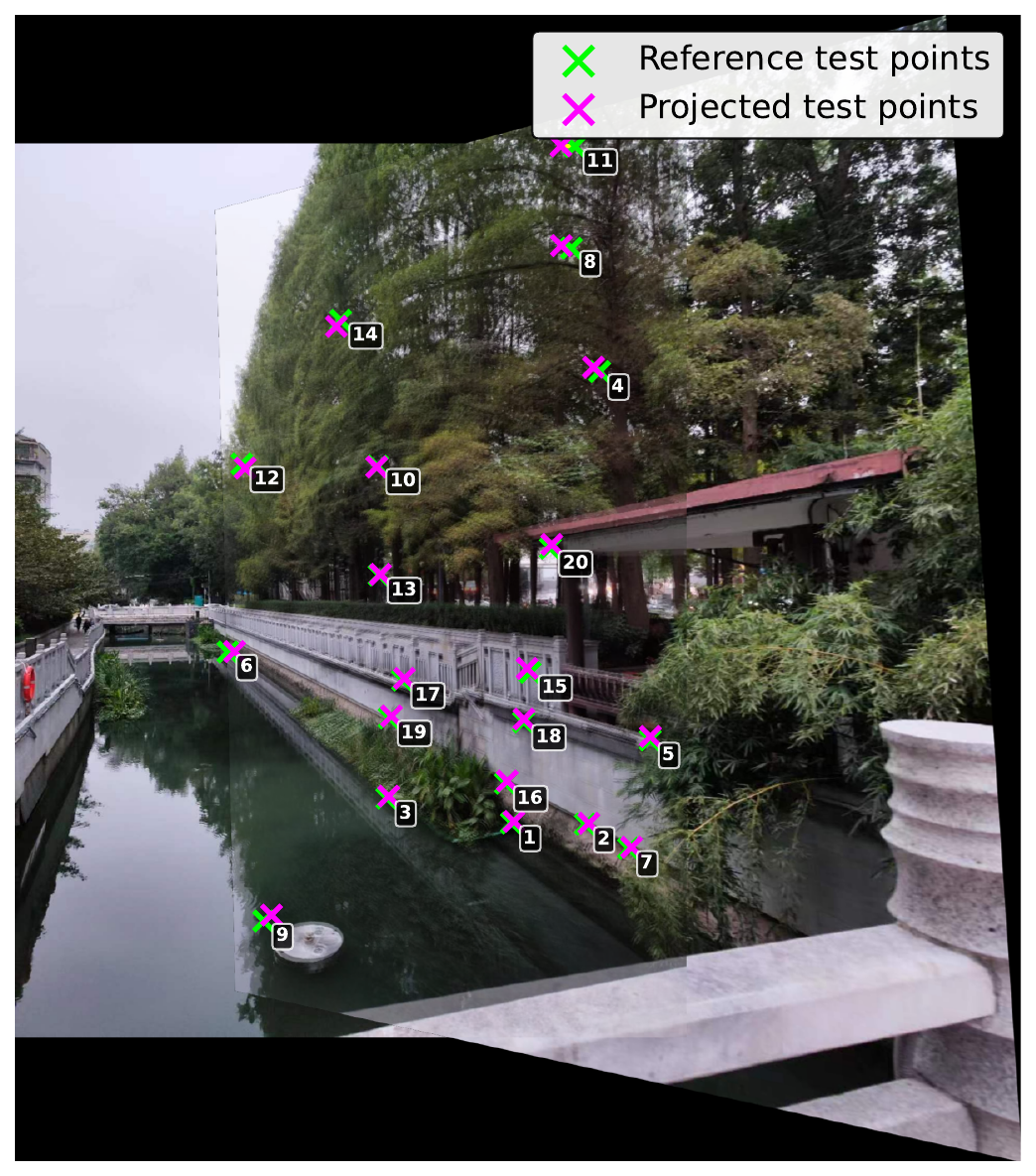}
        \caption{Iterative Bayesian approach (RMSE$=11.44\mathrm{px}$).}
    \end{subfigure}
    \caption{Comparison of a OBJ-GSP-03\_river panorama.}
\label{appfig:river}
\end{figure*}

The numerical transformation parameters for the  OBJ-GSP-03\_river images from the StitchBench \cite{cai2025stitchbench} dataset are

\begin{align}
\mathbf{R}^\mathrm{DLT} = &\begin{bmatrix}
        0.61 & 0.00 & 383.19 \\
        -0.25 & 0.79 & 148.53 \\
        0.00 & 0.00 & 1.00  
    \end{bmatrix}~~, \\
\mathbf{R}_\mathrm{m}^\mathrm{Bayes} = &\begin{bmatrix}
        0.66 & 0.01 & 381.54 \\
        -0.23 &  0.83 & 126.80 \\
        0.00 & 0.00 & 1.00  
\end{bmatrix} \\ \pm &\begin{bmatrix}
        0.02 & 0.01 & 14.84 \\
        0.02 & 0.01 & 14.84 \\
        0.01 & 0.00 & 4.69
\end{bmatrix}.
\end{align}
The resulting panoramas are shown in \cref{appfig:river}.

\begin{table}[!t]
    \centering
    \caption{Parameters of the image stitching examples.}
    \label{tab:imgstich_parameters}
    \small
    \setlength{\tabcolsep}{3pt}

    \begin{tabularx}{\columnwidth}{
        l
        c
        c
        >{\centering\arraybackslash}X
        c
        c
    }
        \toprule
        Image &
        $\sigma_n\,(\mathrm{px})$ &
        Method &
        $\hat{w}^\mathrm{init}$ &
        Prior &
        RMSE $(\mathrm{px})$ \\
        \midrule

        Skyline
            & 10.00 & DLT
            & -- & -- & 47.42 \\

        Skyline
            & 10.00 & Bayes
            & $\hat{w}^\mathrm{init}_\mathrm{Bayes}$
            & 0.10 & 8.92 \\

        v\_graffiti
            & 5.00 & DLT
            & -- & -- & 16.95 \\

        v\_graffiti
            & 5.00 & Bayes
            & $\hat{w}^\mathrm{init}_\mathrm{DLT}$
            & None & 14.11 \\

        River
            & 10.00 & DLT
            & -- & -- & 17.03 \\

        River
            & 10.00 & Bayes
            & $\hat{w}^\mathrm{init}_\mathrm{DLT}$
            & None & 11.44 \\

        Restaurant
            & 5.00 & DLT
            & -- & -- & 277.45 \\

        Restaurant
            & 5.00 & Bayes
            & $\hat{w}^\mathrm{init}_\mathrm{Bayes}$
            & 10.00 & 235.88 \\

        Palazzo
            & 5.00 & DLT
            & -- & -- & 14.48 \\

        Palazzo
            & 5.00 & Bayes
            & $\hat{w}^\mathrm{init}_\mathrm{DLT}$
            & None & 11.48 \\

        \bottomrule
    \end{tabularx}
\end{table}

\subsection{Aerial/Restaurant}
\begin{figure*}
  \centering
   \begin{subfigure}[t]{0.49\linewidth}
        \centering
        \includegraphics[width=\linewidth]{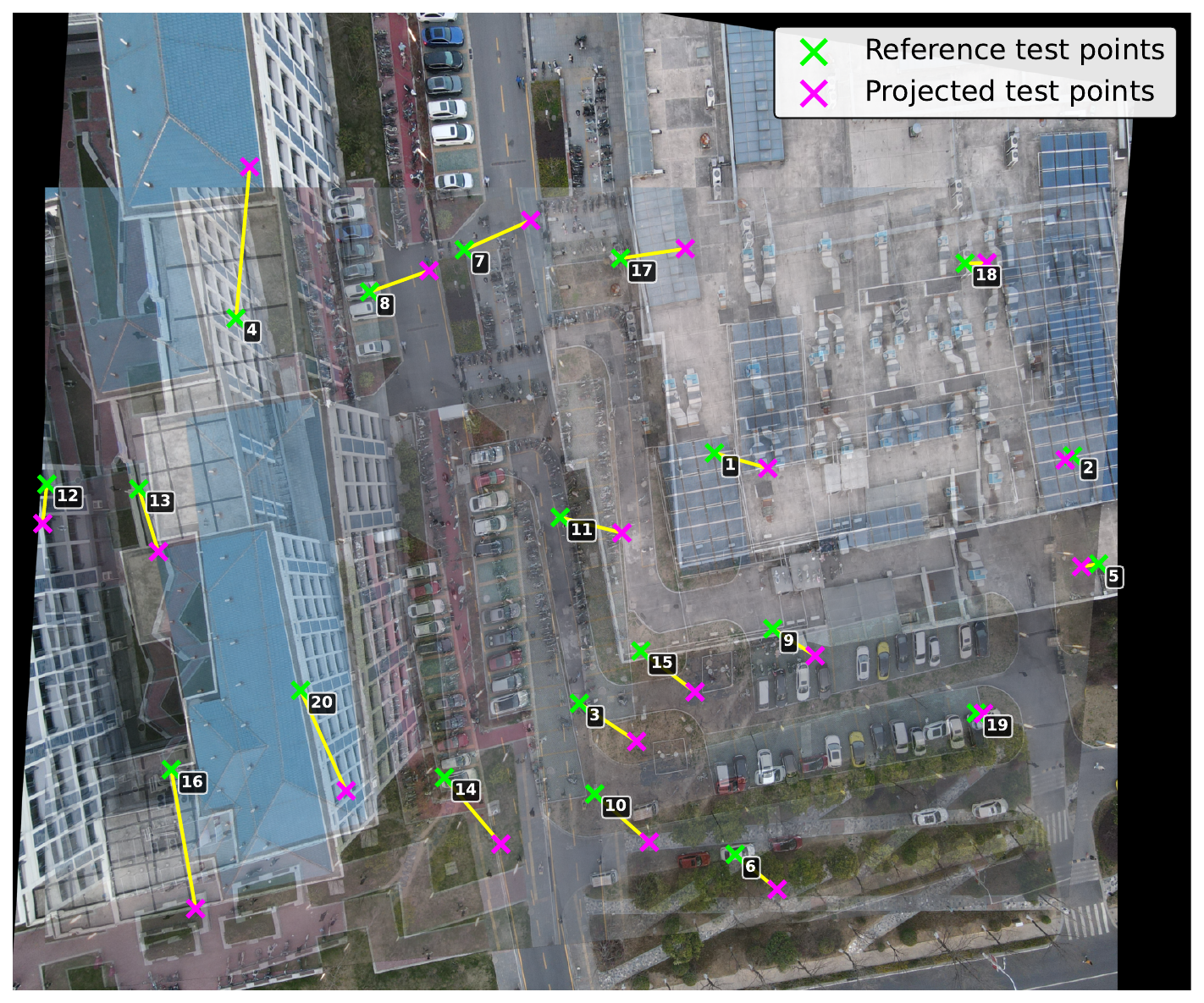}
        \caption{DLT estimation (RMSE$=277.45\,\mathrm{px}$).}
    \end{subfigure}
     \begin{subfigure}[t]{0.49\linewidth}
        \centering
        \includegraphics[width=\linewidth]{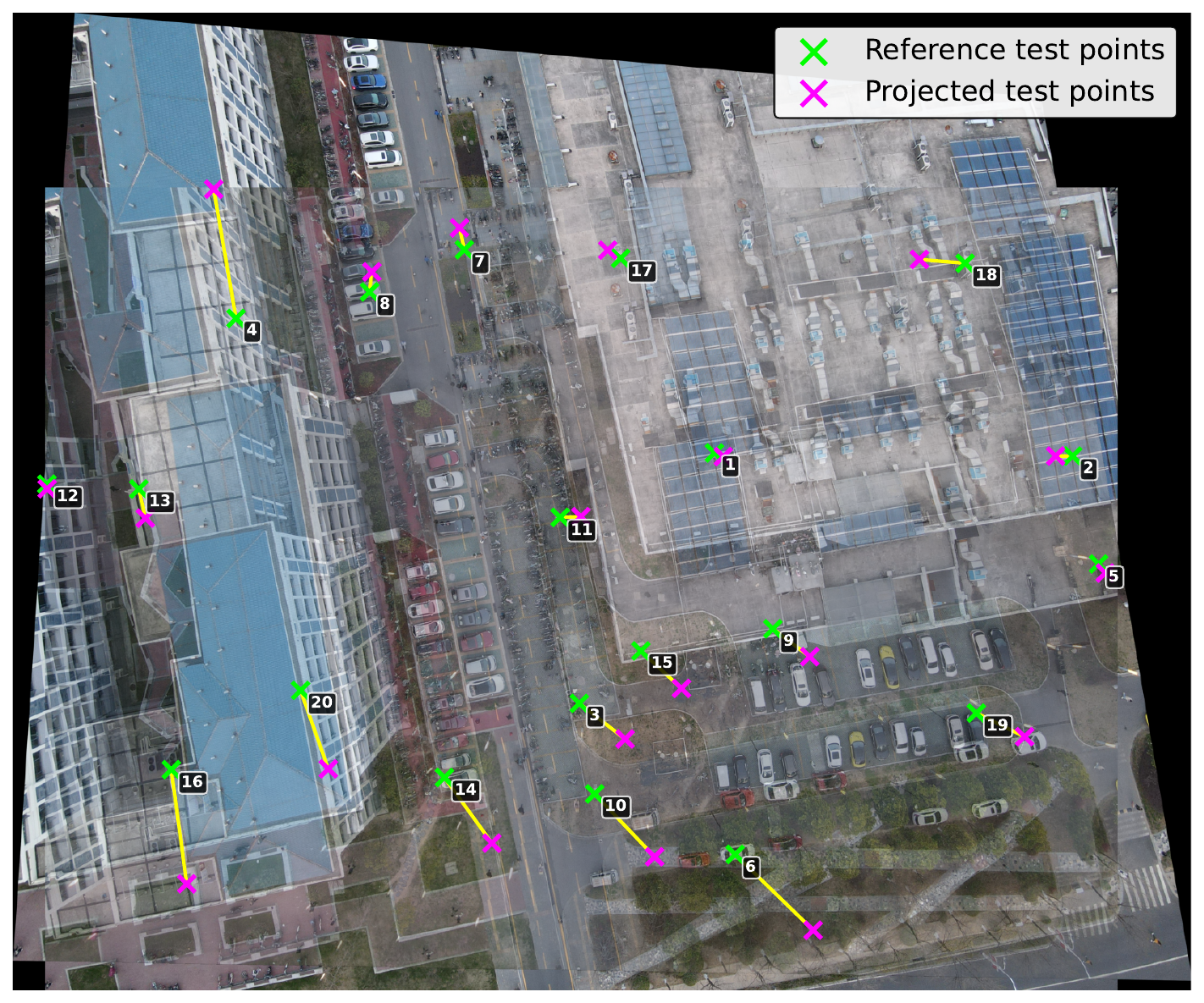}
        \caption{Iterative Bayesian approach (RMSE$=235.88,\mathrm{px}$).}
    \end{subfigure}
    \caption{Comparison of a Aerial/Restaurant panorama.}
    \label{appfig:restaurant}
\end{figure*}

The numerical transformation parameters for the  Aerial/Restaurant images\footnote{Images DJI\_0036.jpg and DJI\_0037.jpg.} from the StitchBench \cite{cai2025stitchbench} dataset are
\begin{align}
\mathbf{R}^\mathrm{DLT} = &\begin{bmatrix}
    1.33 & -0.08 & 108.94 \\ 
    0.13 & 1.36 & -995.27 \\
    0.00 & 0.00 & 1.00
    \end{bmatrix}~~, \\
\mathbf{R}_\mathrm{m}^\mathrm{Bayes} = &\begin{bmatrix}
        0.97 & -0.07 & 110.52 \\
        0.07 & 1.00 &  -649.48 \\
        0.00 & 0.00 & 1.00  
\end{bmatrix} \\ \pm &\begin{bmatrix}
        0.00 & 0.00 & 2.80 \\
        0.00 & 0.00 & 2.80 \\
        0.00 & 0.00 & 2.04
\end{bmatrix}.
\end{align}
The resulting panoramas are shown in \cref{appfig:restaurant}.

\subsection{NISwGSP-06\_PalazzoPubblico}
\label{app:palazzo}
\begin{figure*}
  \centering
   \begin{subfigure}[t]{0.49\linewidth}
        \centering
        \includegraphics[width=\linewidth]{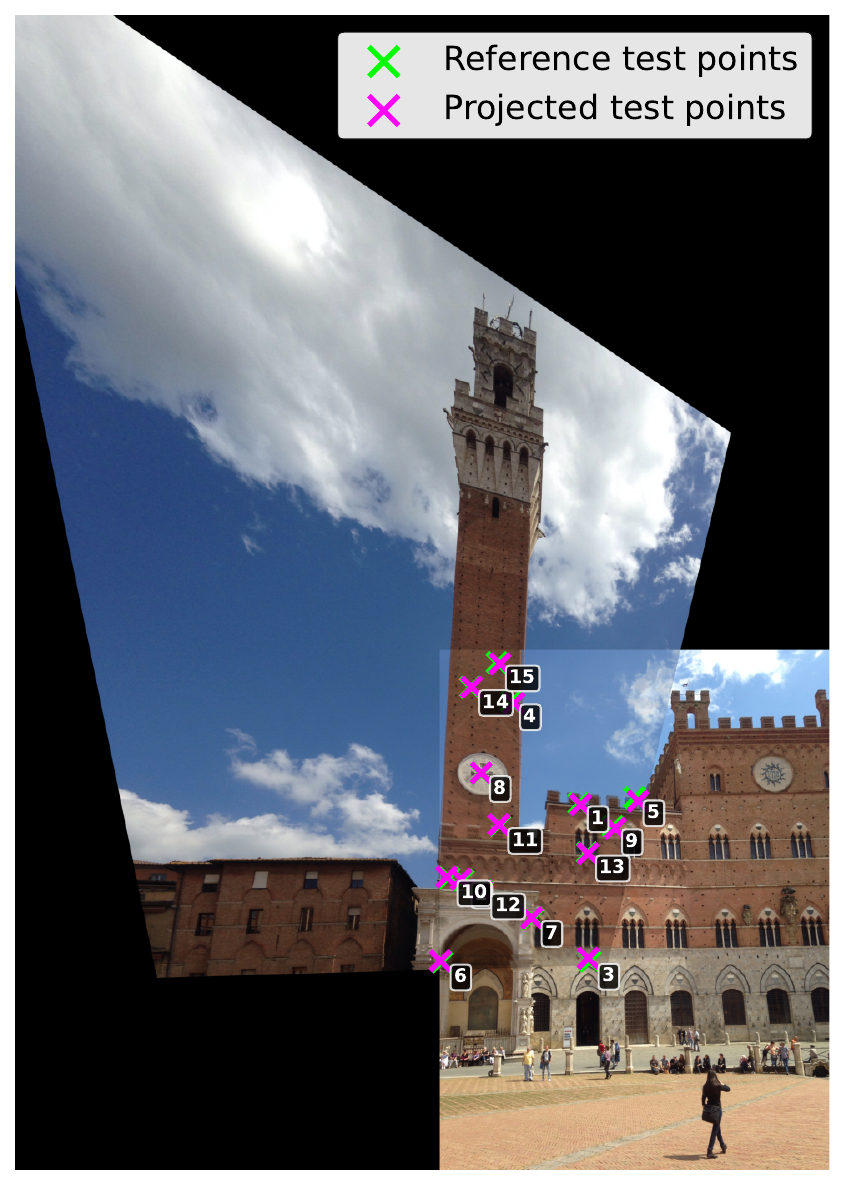}
        \caption{DLT estimation (RMSE$=14.48\,\mathrm{px}$).}
    \end{subfigure}
     \begin{subfigure}[t]{0.49\linewidth}
        \centering
        \includegraphics[width=\linewidth]{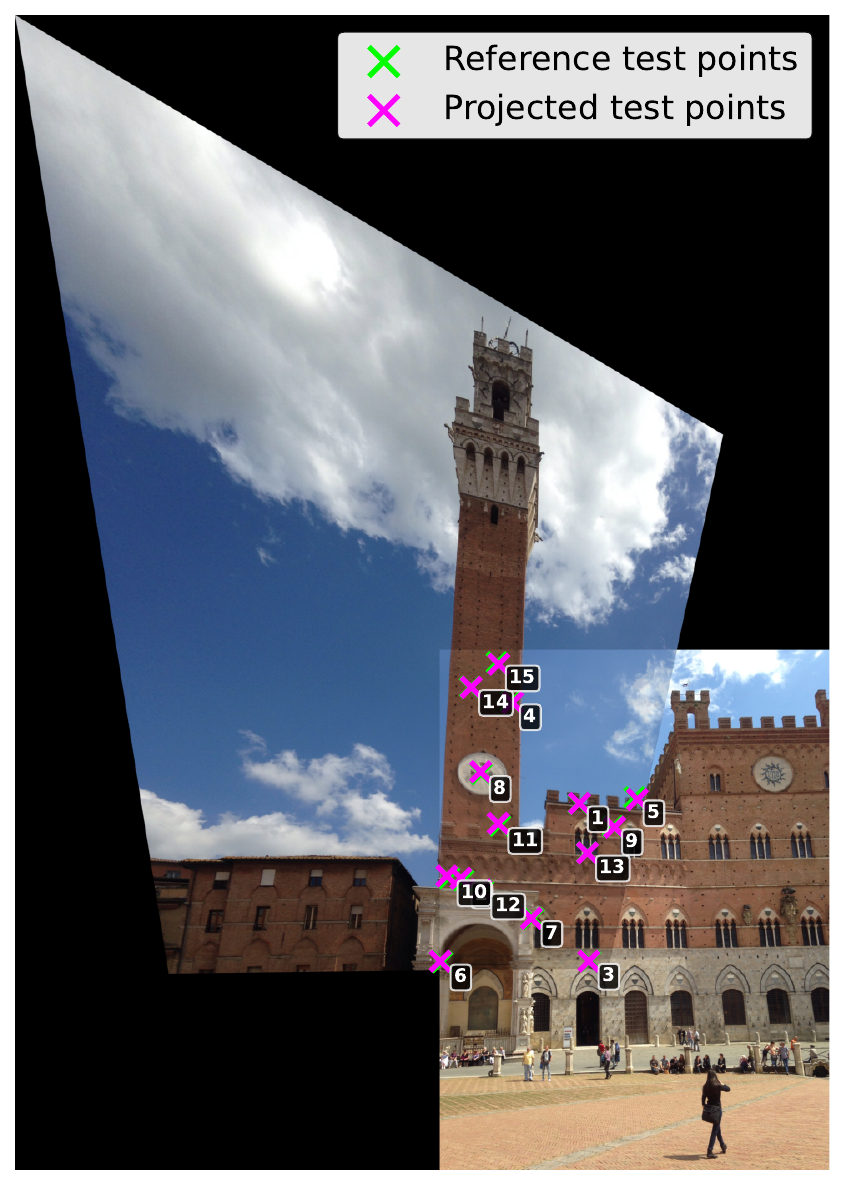}
        \caption{Iterative Bayesian approach (RMSE$= 11.48$ px).}
    \end{subfigure}
    \caption{Comparison of a NISwGSP-06\_PalazzoPubblico panorama.}
    \label{appfig:palazzopublico}
\end{figure*}

The numerical transformation parameters for the NISwGSP-06\_PalazzoPubblico \cite{Chen:2016:NIS} images\footnote{Images image02.jpg and image04.jpg.} from the StitchBench \cite{cai2025stitchbench} dataset collection are
\begin{align}
\mathbf{R}^\mathrm{DLT} = &\begin{bmatrix}
    2.81 & -0.15 & -3145.30 \\ 
    0.78 & 2.74 & -4715.48 \\
    0.00 & 0.00 & 1.00
    \end{bmatrix}~~,\\ 
    \mathbf{R}_\mathrm{m}^\mathrm{Bayes}=
    &\begin{bmatrix}
     2.39 & -0.13 & -2651.55 \\ 
     0.63 & 2.35 & -3969.74 \\
     0.00 & 0.00 & 1.00   
    \end{bmatrix} ~~,\\
    \equiv &\begin{bmatrix}
     1.30 & -0.07 & -1440.42 \\ 
     0.34 & 1.27 & -2156.52 \\
     0.00 & 0.00 & 0.54 
\end{bmatrix} \\ \pm &\begin{bmatrix}
    0.01 & 0.01 & 12.89 \\ 
     0.01 & 0.01 & 12.89 \\
     0.00 & 0.00 & 5.77 
\end{bmatrix}.
\end{align}
The resulting panoramas are shown in \cref{appfig:palazzopublico}.

\clearpage

\end{document}